\pdfoutput=1
\documentclass[11pt]{article}

\usepackage[final]{acl}
\usepackage{times}
 \usepackage{microtype}

\usepackage{booktabs}  
\usepackage{graphicx}
\usepackage{arydshln}

\usepackage{multirow}
\usepackage{float}
\usepackage{amsmath}
\usepackage{amssymb}
\usepackage{CJKutf8}
\DeclareMathOperator*{\argmax}{arg\,max}
\DeclareMathOperator*{\argmin}{arg\,min}

\newlength{\maxwidth}
\newcommand{\algalign}[2]
{\makebox[\maxwidth][r]{$#1{}$}${}#2$}

\newcommand{\ourmethod}{HydraOpt}

\usepackage{subcaption}

\usepackage{xcolor}
\definecolor{darkred}{rgb}{0.6,0.0,0.0}
\definecolor{darkgreen}{rgb}{0,0.50,0}
\definecolor{lightblue}{rgb}{0.0,0.42,0.91}
\definecolor{orange}{rgb}{0.99,0.48,0.13}
\definecolor{grass}{rgb}{0.18,0.80,0.18}
\definecolor{pink}{rgb}{0.97,0.15,0.45}

\usepackage{listings}

\lstdefinestyle{colored}{ %
  basicstyle=\ttfamily\footnotesize,
  backgroundcolor=\color{white},
  commentstyle=\color{green}\itshape,
  keywordstyle=\color{blue}\bfseries\itshape,
  stringstyle=\color{red},
}
\lstdefinelanguage{PythonPlus}[]{Python}{
  morekeywords=[1]{,as,assert,nonlocal,with,yield,self,True,False,None,} 
  morekeywords=[2]{,__init__,__add__,__mul__,__div__,__sub__,__call__,__getitem__,__setitem__,__eq__,__ne__,__nonzero__,__rmul__,__radd__,__repr__,__str__,__get__,__truediv__,__pow__,__name__,__future__,__all__,}, 
  morekeywords=[3]{,object,type,isinstance,copy,deepcopy,zip,enumerate,reversed,list,set,len,dict,tuple,range,xrange,append,execfile,real,imag,reduce,str,repr,}, 
  morekeywords=[4]{,Exception,NameError,IndexError,SyntaxError,TypeError,ValueError,OverflowError,ZeroDivisionError,}, 
  morekeywords=[5]{,ode,fsolve,sqrt,exp,sin,cos,arctan,arctan2,arccos,pi, array,norm,solve,dot,arange,isscalar,max,sum,flatten,shape,reshape,find,any,all,abs,plot,linspace,legend,quad,polyval,polyfit,hstack,concatenate,vstack,column_stack,empty,zeros,ones,rand,vander,grid,pcolor,eig,eigs,eigvals,svd,qr,tan,det,logspace,roll,min,mean,cumsum,cumprod,diff,vectorize,lstsq,cla,eye,xlabel,ylabel,squeeze,}, 
}
\lstdefinelanguage{PyBrIM}[]{PythonPlus}{
  emph={d,E,a,Fc28,Fy,Fu,D,des,supplier,Material,Rectangle,PyElmt},
}
\lstdefinestyle{colorEX}{
  basicstyle=\ttfamily\footnotesize,
  backgroundcolor=\color{white},
  commentstyle=\color{darkgreen}\slshape,
  keywordstyle=\color{blue}\bfseries\itshape,
  keywordstyle=[2]\color{blue}\bfseries,
  keywordstyle=[3]\color{grass},
  keywordstyle=[4]\color{red},
  keywordstyle=[5]\color{orange},
  stringstyle=\color{darkred},
  emphstyle=\color{pink}\underbar,
}

\usepackage{algorithm}
\usepackage[noend]{algpseudocode}

\newcommand{\keypoint}[1]{\noindent \textbf{#1}\quad}

\makeatletter
\def\eg{\textit{e.g.}} 
\def\ie{\textit{i.e.}}

\makeatother

\usepackage[english]{babel} %

\usepackage{tikz}
\usetikzlibrary{decorations.pathreplacing,calc}

\newcommand{\tikzmark}[2][-3pt]{\tikz[remember picture, overlay, baseline=-0.5ex]\node[#1](#2){};}

\tikzset{brace/.style={decorate, decoration={brace}},
 brace mirrored/.style={decorate, decoration={brace,mirror}},
}

\newcounter{brace}
\newcommand{\drawbrace}[3][brace]{%
 \refstepcounter{brace}
 \tikz[remember picture, overlay]\draw[#1] (#2.center)--(#3.center)node[pos=0.5, xshift=-10pt, yshift=6ex, name=brace-\thebrace]{};
}
\newcommand{\drawbraceshort}[3][brace]{%
 \refstepcounter{brace}
 \tikz[remember picture, overlay]\draw[#1] (#2.center)--(#3.center)node[pos=0.5, xshift=-10pt, yshift=3ex, name=brace-\thebrace]{};
}

\newcounter{arrow}
\newcommand{\annote}[3][]{%
 \tikz[remember picture, overlay]\node[#1] at (#2) {#3};
}

\title{\ourmethod: Navigating the Efficiency-Performance Trade-off of Adapter Merging}

\author{
 \textbf{Taha Ceritli\textsuperscript{1}},
 \textbf{Ondrej Bohdal\textsuperscript{1}},
 \textbf{Mete Ozay\textsuperscript{1}},
 \\
 \textbf{Jijoong Moon\textsuperscript{2}},
 \textbf{Kyeng-Hun Lee\textsuperscript{2}},
 \textbf{Hyeonmok Ko\textsuperscript{2}},
 \textbf{Umberto Michieli\textsuperscript{1}}
\\
 \textsuperscript{1}Samsung R\&D Institute UK, United Kingdom,
 \textsuperscript{2}Samsung Research, South Korea
\\
 \small{
   \textbf{Correspondence:} \href{mailto:o.bohdal.1@samsung.com}{t.ceritli@samsung.com}
 }
}

\begin{document}

\maketitle
\begin{abstract}
Large language models (LLMs) often leverage adapters, such as low-rank-based adapters, to achieve strong performance on downstream tasks. However, storing a separate adapter for each task significantly increases memory requirements, posing a challenge for resource-constrained environments such as mobile devices. Although model merging techniques can reduce storage costs, they typically result in substantial performance degradation. In this work, we introduce \ourmethod, a new model merging technique that capitalizes on the inherent similarities between the matrices of low-rank adapters. Unlike existing methods that produce a fixed trade-off between storage size and performance, \ourmethod~allows us to navigate this spectrum of efficiency and performance. Our experiments show that \ourmethod~significantly reduces storage size (48\% reduction) compared to storing all adapters, while achieving competitive performance (0.2-1.8\% drop). Furthermore, it outperforms existing merging techniques in terms of performance at the same or slightly worse storage efficiency. 

\end{abstract}

\section{Introduction}
Large language models (LLMs) have become a driving force behind many natural language processing tasks today, including text summarization \cite{liu2023learning}, smart-reply \cite{bastola2023llm} and question-answering \cite{sticha2024qa}. 
While modern LLMs are pre-trained to perform a diverse set of tasks, their performance on specific tasks can be improved by updating its parameters on task-specific datasets. 
However, this fine-tuning process becomes computationally impractical due to the growing size of LLMs, especially for resource-constrained environments such as mobile devices.

\begin{figure}[t]
\centering
    \includegraphics[width=\linewidth]{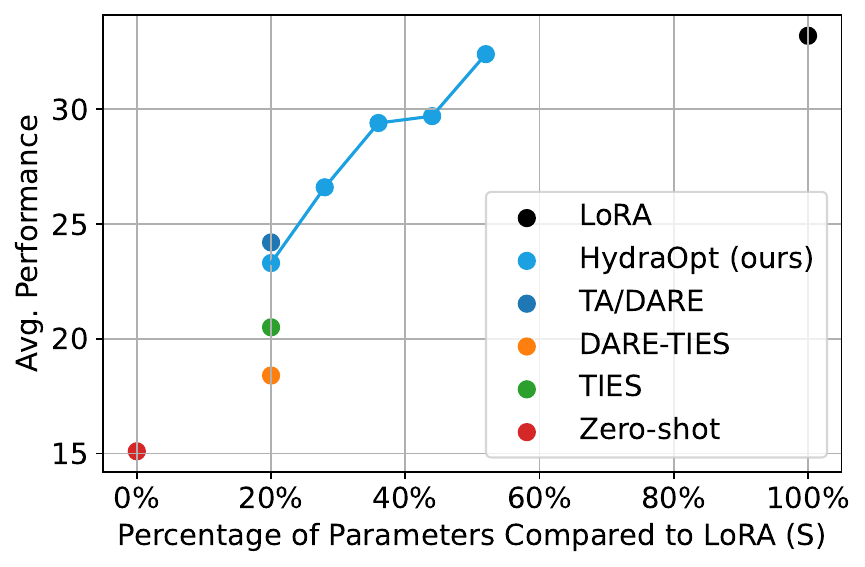}
  \caption{\textbf{Performance and storage efficiency tradeoff.} Average performance over 5 applications and 8 languages. 
  Existing merging techniques reduce storage costs at significant performance drops. Our method performs similarly at the same efficiency level and improves if more storage is available, achieving performance similar to LoRAs.}

  \label{fig:merging_tradeoff}
\end{figure}

One approach to reduce the computational complexity of the fine-tuning process is parameter-efficient fine-tuning (PEFT) \citep{hu2021lora,xu2023parameter,lialin2023scaling}, where only a small set of parameters is updated while keeping the parameters of the LLMs frozen. 
For example, low-rank-based adapters such as LoRA \cite{hu2021lora} and VeRA \cite{kopiczko2023vera} have facilitated the use of LLMs for on-device applications, as one can store separate adapters for different tasks and switch to the corresponding parameters when the user wishes to perform a specific task \cite{gunter2024apple}. However, storing separate adapters becomes costly for on-device settings where the memory is limited. Model merging techniques \citep{wortsman2022model,ilharco2022editing,yadav2024ties,yu2024language} address this issue by combining multiple adapters into one adapter used for all tasks. However, such techniques significantly disrupt the performance with no control. 

In this work, we propose a new model merging method (\ourmethod) that allows controllable efficiency-performance trade-off, unlike existing methods that result in a single storage size and performance.
\ourmethod~achieves competitive performance compared to storing all adapters but with reduced storage size and improves over the performance of existing merging techniques at a slightly worse storage efficiency (Fig.~\ref{fig:merging_tradeoff}).
Our contributions are three-fold:
\begin{itemize}

    \item Building on the similarity between low-rank-based adapters, we introduce a new model merging strategy called \ourmethod~that can navigate the efficiency-performance trade-off of model merging.

    \item We design a comprehensive evaluation framework that consists of 40 tasks derived from 5 applications and 8 languages. We conduct experiments to assess the impact of merging adapters across applications, languages, and tasks.

    \item Our experiments demonstrate that \ourmethod~finds a better trade-off between efficiency and performance across different low-rank-based adapters and LLMs. While maintaining comparable performance to existing model merging methods at similar storage levels, \ourmethod~consistently outperforms them when a modest increase in storage is permitted.

\end{itemize}

\section{Related Work}
\label{sec:related}
\keypoint{Parameter-efficient Fine-tuning (PEFT)} techniques adapt models efficiently via training relatively few parameters, making them especially suitable for fine-tuning large language models \cite{ding2022delta,han2024parameter}.  Low-rank-based adapters \cite{hu2021lora,liu2024dora,kopiczko2023vera,malinovsky2024randomized,ceritli2024}, in particular, have become widely adopted, having small additional storage requirements thanks to their compact size, which makes them suitable for deployment to mobile devices \cite{gunter2024apple}. 
LoRA \cite{hu2021lora} introduces two low-dimensional trainable parameters ${A\in \mathbb{R}^{r\times k}}$ and ${B\in \mathbb{R}^{d\times r}}$ which are used to approximate the weight updates $\Delta W$, \ie, $\Delta W = BA$ where rank $r << \min(d, k)$. Then, the LLM parameters can be updated such that $W = W + \Delta W$. 
Performance of LoRA has been further improved in its many extensions, such as AdaLoRA \cite{zhang2023adalora} and DoRA \cite{liu2024dora}.

Various approaches to improve efficiency of LoRA have also been proposed \cite{kopiczko2023vera,renduchintala2024tied}.
In particular, VeRA \cite{kopiczko2023vera} has become popular for improving storage and parameter efficiency of LoRA, while maintaining competitive performance. 
VeRA introduces the following model update: $\Delta W = \Lambda_b B \Lambda_d A$ where the parameters ${B\in \mathbb{R}^{d\times r}}$ and ${A\in \mathbb{R}^{r\times k}}$ are shared across the layers while the parameters $\Lambda_b \in \mathbb{R}^{d}$ and $\Lambda_d \in \mathbb{R}^{r}$ are defined per each layer. The resulting method reduces the number of trainable parameters, as the layer-specific parameters are defined as vectors rather than matrices.

\keypoint{Model Merging:} Multiple task-specific models can be combined into a single model capable of multi-tasking via a process called model merging. Task Arithmetic \cite{wortsman2022model,ilharco2022editing} represents the simplest option, and it combines the weights of multiple models as a weighted average. Various more advanced techniques have been developed, including  TIES \cite{yadav2024ties} and DARE \cite{yu2024language}. TIES first resets the values of parameters that changed little, then elects the sign in case of conflicts, and merges only sign-aligned parameters. DARE drops part of the weight changes and then rescales the remaining ones accordingly.
Other methods \cite{xiao2024lm,huang2023lorahub,hammoud2024model,shenaj2024lora} use data to improve merging, however, that is beyond the scope of the present work.

\keypoint{On-device LLMs:} 
LLMs typically include billions of parameters, which requires significant resources, such as high-end GPUs,
even for inference only \cite{borzunov2024distributed}. However, in many use cases, it is desirable to perform 
computations locally without transferring data to remote servers \cite{dhar2021ondevice}, for example, when using sensitive data stored on resource-constrained devices. Real-world examples include generating personalized replies or summarizing private conversations, where maintaining data privacy is paramount. As a solution, smaller LLMs (\eg, 1–3 billion parameters) have been developed for on-device deployment. These models utilize model compression strategies paired with a smaller size to support efficient on-device inference. Prominent examples include Llama 3.2 1B \cite{dubey2024llama}, StableLM2 1.6B \cite{bellagente2024stable} and Qwen2.5 1.5B \cite{qwen2,qwen2.5}. Due to their relatively small size, it is standard practice to include single-task adapters on the device to enable the small LLMs to perform the individual tasks, instead of relying on instruction following \cite{gunter2024apple,dong2024survey}. 

\section{Proposed Method}

\subsection{Motivation}

Our work stems from the analysis of the asymmetric behaviour of low-rank adaptation matrices. \citet{zhu2024asymmetry} demonstrate that the $B$ parameters in LoRA exhibit distinct values when fine-tuned across different tasks, while the $A$ parameters remain relatively similar when initialized identically, despite being fine-tuned on diverse tasks. 
Similarly, \citet{tian2024hydralora} observe that when multiple LoRA adapters are trained on separate datasets, the $A$ parameters tend to converge to similar values, whereas the $B$ parameters become more differentiated.

\begin{figure}[!htb]
\centering
    \includegraphics[width=\linewidth]{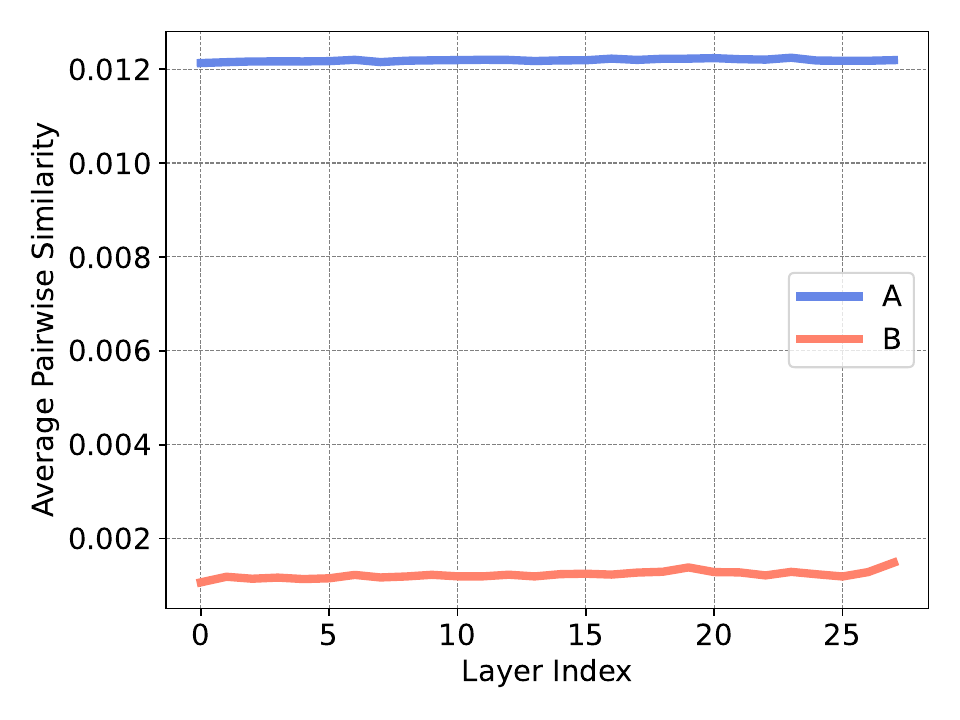}
  \caption{\textbf{Similarity between A and B matrices of LoRAs} measured using Mean Absolute Error on query matrices of Llama-3.2-3B-Instruct fine-tuned on 5 applications in English.}
  \label{fig:similarities_avg}
\end{figure}

We observe similar patterns when fine-tuning Llama-3.2-3B-Instruct on five distinct text generation tasks in English. Fig.~\ref{fig:similarities_avg} illustrates the similarity between these LoRA adapters computed via Mean Absolute Error. The plot indicates that the $A$ parameters are more similar to each other compared to the $B$ parameters. 
We report in Fig.~\ref{fig:similarities_avg_cca} an analysis using Canonical Correlation Analysis (CCA), as in \citet{zhu2024asymmetry}. We further confirm these results with t-SNE visualizations in Fig.~\ref{fig:tsne_results} following the approach of \citet{tian2024hydralora}.

These findings suggest that the $A$ parameters capture cross-domain commonalities, while the $B$ parameters adapt to task-specific knowledge. This behavior may stem from the initialization schemes for low-rank-based adapters, where $B$ is typically initialized as a zero matrix, while $A$ is sampled from a Gaussian distribution. These behaviors can also be observed in VeRA as $\Lambda_b$ is initialized as zero vector and $\Lambda_d$ is sampled from a Gaussian distributions.

\subsection{Our Method: \ourmethod}

\begin{figure}[t]
    \includegraphics[width=\linewidth]{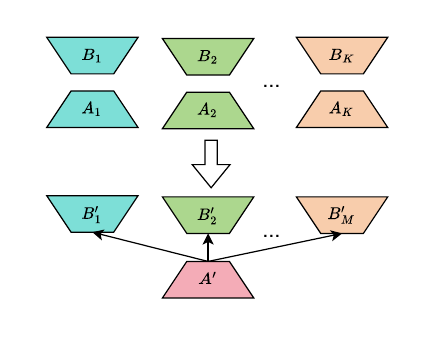}
  \caption{\textbf{An overview of \ourmethod.} We approximate $K$ sets of LoRA parameters by learning a shared $A^\prime$ parameter and a set of task-specific parameters $\{B_i^\prime\}_{i=1}^M$.}
  \label{fig:overview}
\end{figure}

We propose \ourmethod~for merging a set of low-rank-based adapters parameters (\eg~LoRA) $\{B_i, A_i\}_{i=1}^K$. As shown in Fig.~\ref{fig:overview}, \ourmethod~approximates the given set of parameters by learning a shared $A^\prime$ parameter and a set of $B^\prime$ parameters $\{B_i^\prime\}_{i=1}^M$. The approximation to the original model updates is driven by the following loss function:

\begin{equation}
    \ell = \sum_{i=1}^K f\left(B_i A_i, \sum_{j=1}^M \sigma(\mathbf{C}_i^\prime / T) (j) B_j^\prime A^\prime\right),
\label{eq:loss_function_1}
\end{equation}
where $f$ is a distance function that measures the similarity between two model updates $\Delta W_i := B_i A_i$ and $\Delta W_j^\prime := B_j^\prime A^\prime$. Here, $\sigma$ denotes the softmax function with the temperature term $T$, and $\mathbf{C}_i^\prime \in \mathbb{R}^{M}$ is a trainable vector of coefficients with the coefficient $C_{i,j}^\prime$ representing how likely it is to use $B_j^\prime$ for the $i^{th}$ task. The softmax function approximates categorical one-hot encoded vectors for small values, hence guiding the model to use mostly one $B_j^{\prime}$ parameter for a given task. 

Given the sparsity of adapter parameters, we choose Mean Absolute Error as the distance function $f$ to induce sparsity \cite{bach2012optimization}. We then calculate the gradients of the objective function Eq.~\eqref{eq:loss_function_1} to update the parameters $A^\prime, \{B_i^\prime\}_{i=1}^M, \{\mathbf{C}_i^\prime\}_{i=1}^K$ using an iterative optimization algorithm. For instance, the update rule for $A^\prime$ using Gradient Descent becomes $A^\prime = A^\prime - \eta \nabla \ell$ where $\eta$ is the learning rate and $\ell$ is the loss.

The coefficients $\mathbf{C}_i^\prime$ are initialized using a Gaussian distribution and updated during training. We remark that this only brings a minimal increase in memory footprint during training, since the number of coefficients is much smaller than the number of LoRA parameters. Moreover, these coefficients are discarded once the training is over, after we associate each task with a $B^\prime$ parameter. Therefore, the inference stage is unaltered.

\ourmethod~allows us to walk the efficiency-performance trade-off of model merging. In the most aggressive parameter sharing scheme, \ourmethod~constructs one set of LoRA parameters $\{B^\prime, A^\prime\}$, which causes performance drops due to the reduced flexibility of the approximation similarly to existing model merging techniques. However, by adjusting the level of parameter sharing, we obtain both efficient and accurate model merging solutions. 

In the special case of $K=M$, we omit the coefficients $\mathbf{C}_i^\prime$ by learning a separate $B^\prime$ parameter for each task. Specifically, we modify the loss function in Eq.~\eqref{eq:loss_function_1} to be:
\begin{equation}
    \ell = \sum_{i=1}^K f(B_i A_i, B_i^\prime A^\prime).
\label{eq:loss_function_2}
\end{equation}

The reduction in parameter size using \ourmethod~depends on the number of LoRAs to merge and the size of LoRA parameters. For one layer of an LLM, the number of parameters required by LoRA becomes $K \times r \times (d+k)$ for $K$ tasks, whereas \ourmethod~requires $M \times r \times d + r\times k$ parameters where $M$ denotes the number of $B^\prime$ parameters. Assuming that the $A$ and $B$ parameters are the same size, the total number of parameters reduces to 60\% when merging 5 pairs of LoRA parameters (Fig.~\ref{fig:parameter_size_reduction} in Appendix \ref{sec:additional_results}). The reduction rate asymptotically reaches 50\% as the number of LoRAs increases, while it can be greater than 50\% if the $A$ is larger than $B$.

Notice that the objective functions in Eq.~\eqref{eq:loss_function_1}-\eqref{eq:loss_function_2} do not utilize any external task-specific samples. Instead, they treat the given set of LoRA parameters $\{B_i, A_i\}_{i=1}^{K}$ as the target pseudo-labels and updates the trainable \ourmethod~parameters 
$A^\prime, \{B_i^\prime\}_{i=1}^M, \{\textbf{C}_i^\prime\}_{i=1}^K$ 
during the training. Therefore, our technique is classified as a data-free model merging method.
Algorithm~\ref{alg:hydra_merging_alg} provides a brief description of \ourmethod.

\begin{algorithm}[t]
  \caption{\ourmethod}
  \label{alg:hydra_merging_alg}
    \begin{algorithmic}[1]
    \Require Adapter parameters $\{A_i, B_i\}_{i=1}^K$, target number of $B^\prime$ parameters $M$, number of epochs $E$, temperature $T$, optimizer 
    \State{Initialize $A^\prime, \{B_i^\prime\}_{i=1}^M, \{\mathbf{C}_i^\prime\}_{i=1}^K$} 
    \For {$e$ in $E$}
        \If{$K \neq M$}    \State Calculate the loss $\ell$ using Eq.~\eqref{eq:loss_function_1}
        \Else    \State{Calculate the loss $\ell$ using Eq.~\eqref{eq:loss_function_2}}
        \EndIf
        \State Parameter update using the optimizer:
        
        \settowidth{\maxwidth}{$\{B_i^\prime\}_{i=1}^M$}
        \State \algalign{A^\prime}{\gets \underset{A^\prime} \argmin \ \ell}
        
        \State \algalign{\{B_i^\prime\}_{i=1}^M}{\gets \underset{\{B_i^\prime\}_{i=1}^M}\argmin \ \ell}
        
        \State \algalign{\{\mathbf{C}_i^\prime\}_{i=1}^K}{\gets \underset{\{\mathbf{C}_i^\prime\}_{i=1}^K}\argmin\ \ell }

    \EndFor
        \State {Use $A'$ for all tasks}
        \For {$i$-th task in ${1, 2, ..., K}$}
            \If{$K \neq M$}     \State{$j \gets \underset{j \in \{1, 2, ..., M\}} \argmax{C_{i,j}^\prime}$}
                \State{Use $B_j^\prime$ rather than $B_i$} 
            \Else     \State{Use $B_i^\prime$ rather than $B_i$}
            \EndIf
        \EndFor
    \end{algorithmic}
\end{algorithm}

\section{Experiments}
In this section, we describe our experimental setup (Sec.~\ref{sec:setup}) and discuss the main results (Sec.~\ref{sec:main_results}) and the ablation studies (Sec.~\ref{sec:ablation_results}).

\subsection{Setup}
\label{sec:setup}
\noindent\textbf{Tasks:} We conduct experiments on 40 downstream tasks in total, each of which is a text generation application in a specific language.
We tackle 5 applications.
(i) Grammar Correction (GC): to generate the correct form of a given input containing grammar errors.
(ii) Smart Reply (SR): to generate a response to a given textual message.
(iii) Text Summarization (TS): to generate a shorter version of a given sentence.
(iv) Tone Adjustment (TA): to re-write a given text in a specific style.
(iv) Question Answering (QA): to answer a given question. Each application is considered in 8 different languages: EN, DE, ES, FR, IT, JA, KO, ZH\footnote{Abbreviated for English, German, Spanish, French, Italian, Japanese, Korean, Chinese, respectively}.

\noindent\textbf{Datasets:} 
We use Cambridge English Write \& Improve (W\&I) \cite{bryant2019bea} for GC, Persona-Chat Synthetic \cite{jandaghi2023faithful} for SR, SAMSum \cite{gliwa2019samsum} for TS, Sound Natural \cite{einolghozati2020sound} rephrased using the fine-tuned RedPajama-INCITE-Base-3B-v1 model \cite{utsav2023tone} for TA, and SQuAD \cite{rajpurkar-etal-2016-squad} for QA. 
As these datasets are collected in English, we utilize machine-translation for the remaining languages.
Specifically, we use OPUS-MT \cite{TiedemannThottingal:EAMT2020} for translation to French, German, Italian and Spanish, and M2M100 \cite{fan2021beyond} for translation to Chinese, Japanese and Korean. However, the translation process fixes the grammar errors in the input text (as also mentioned in \citealp{luhtaru2024no}). Therefore, we used datasets collected in the original languages for Grammar Correction, namely ECSpell \cite{lv2023general} for Chinese, Merlin \cite{boyd2014merlin} for Italian, and GitHub Typo Corpus \cite{hagiwara-mita-2020-github} for the remaining languages. Table~\ref{table:datasets} presents a summary of the employed datasets, including their links and number of samples.
Table~\ref{tab:prompts} lists the prompts that we have used for each case.

\noindent\textbf{Evaluation Metrics:} Following the literature, we report F05 ($\uparrow$) \cite{bryant2017automatic} for GC\footnote{We use ChERRANT \cite{zhang-etal-2022-mucgec} for Chinese.}, Weighted Rouge ($\uparrow$) for SR, RougeL ($\uparrow$) for TS and TA, and F1 ($\uparrow$) for QA. Given the page limit, we report the average of these metrics as aggregated overview when needed,  with individual results reported in Appendix \ref{sec:additional_results} for completeness. 
Additionally, we report the storage (S, \%, $\downarrow$) as the percentage of the parameters compared to storing all individual adapters.

\noindent\textbf{Models:} We use Llama-3.2-1B-Instruct \cite{grattafiori2024llama} for our main experiments due to its size suitable for on-device deployment. However, we also present analyses with different model sizes (2B, 3B, and 3.5B) and architectures (Gemma2, \citealp{team2024gemma}, and Phi-3, \citealp{abdin2024phi}) to prove the generalization of our method.

\noindent\textbf{Baseline Methods:} For fair comparison, we compare our method with existing data-free model merging techniques such as Task Arithmetic (TA, \citealt{ilharco2022editing}), TIES \citep{yadav2024ties}, DARE \citep{yu2024language}, and DARE-TIES\footnote{See the dare\_ties function at \url{https://github.com/huggingface/peft/blob/main/src/peft/utils/merge_utils.py} [Accessed on 17 May 2025]}.

\noindent\textbf{Implementation Details:} We set the LoRA rank to 32, $\alpha$ to 128 and dropout to 0.05 throughout the experiments. We utilize the AdamW optimizer with a learning rate of 5e-5 and a batch size of 3. Please see Appendix \ref{sec:datasets} for further details.

\subsection{Main Results}
\label{sec:main_results}
\begin{table}[t]
\small
\setlength{\tabcolsep}{1.7pt}
\caption{\textbf{Performance on 5 English applications using Llama-3.2-1B-Instruct LoRA-finetuned.} 
S represents the percentage of the parameters compared to storing 5 LoRAs.}
\label{table:table1}
\begin{center}
\begin{tabular}{lc|ccccc|c}
\toprule
Method & S (\%) & GC & SR & TS & TA & QA & Avg\\
\midrule
Zero-shot & 0 & 13.1 & 5.1 & 23.4 & 27.6 & 15.8 & 17.0 \\
LoRA & 100 & 35.1 & 23.0 & 38.2 & 58.1 & 61.5 & 43.2 \\
\hdashline
TA & 20 & 25.9 & 11.4 & 32.3 & 51.7 & 28.7 & 30.0 \\
TIES & 20 & 25.1 & 13.2 & 31.1 & 51.7 & 26.9 & 29.6 \\
DARE & 20 & 21.6 & 7.4 & 27.1 & 39.0 & 19.7 & 23.0 \\
DARE-TIES & 20 & 22.5 & 7.9 & 27.4 & 44.0 & 20.9 & 24.5 \\
\ourmethod (M=1) & 20 & 26.3 & 9.0 & 32.3 & 50.4 & 27.6 & 29.1 \\
\ourmethod (M=2) & 28 & 26.9 & 17.6 & 31.7 & 53.0 & 28.2 & 31.5 \\
\ourmethod (M=5) & 52 & 33.1 & 21.7 & 37.1 & 57.1 & 59.9 & 41.8 \\
\bottomrule
\end{tabular}
\end{center}
\end{table}

\noindent\textbf{Merging 5 LoRA adapters in a language:} We first consider a typical model merging scenario where multiple LoRA adapters are obtained by fine-tuning the same model (Llama-3.2-1B-Instruct in this case) across different applications. Table~\ref{table:table1} presents a comparison of the methods in terms of storage efficiency and performance. 
The highest-performing state-of-the-art method is TA that reduces average performance by 13.2\% at the storage occupancy of 20\%. 

\ourmethod~exhibits similar performance at the same storage and starts outperforming TA as the extent of storage size reduction is sacrificed. Specifically, we obtain a 1.5\% gain over TA in average performance when an additional 8\% storage size is used. If more storage is available, we can approach the LoRA upper bound: namely, \ourmethod(M=5) is only 1.4\% lower than LoRA in terms of average performance, while saving $\sim$50\% storage.

We note that \ourmethod~introduces additional runtime; however, the merging operation is still reasonably fast with relatively small GPU memory overhead (please see Table \ref{table:runtime_memory} in
Appendix \ref{sec:additional_results} for a comparison of the methods in terms of runtime and memory).

\begin{figure}[t]
\centering
    \includegraphics[width=\linewidth]{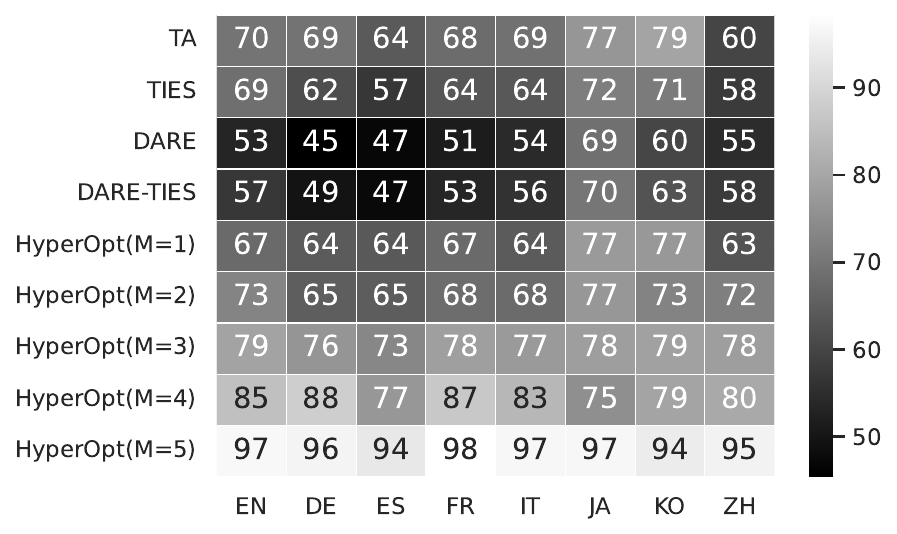}
  \caption{\textbf{Average performance on 5 applications in different languages using Llama-3.2-1B-Instruct.} In this figure, we report the relative average score compared to LoRA.}
  \label{fig:avg_performance_lang}
\end{figure}

\noindent\textbf{Merging 5 LoRA adapters (multiple languages):}
We performed the same 5-way merging experiment in multiple languages and
we observe similar improvements across the board as shown in Fig.~\ref{fig:avg_performance_lang}.

\noindent\textbf{Different LLMs:} Next, we test the generalization of our approach across different LLM sizes and architectures (Llama 1B, Llama 3B, Gemma 2B, Phi3.5B). In  Table~\ref{table:table2}, we observe a consistent trend across model types and sizes. 

\ourmethod (M=5) consistently improves average performance over the baselines at a small storage size cost.
\begin{table}[h]
\small
\setlength{\tabcolsep}{4.5pt}

\caption{\textbf{Average performance on 5 English applications for different LoRA-finetuned LLMs.} We use Llama-3.2-1B-Instruct (L1B), Llama-3.2-3B-Instruct (L3B), Gemma-2-2B-it (G2B), and Phi-3.5-mini-instruct (P3.5B), and report the average of individual metrics for each model.}
\label{table:table2}
\begin{center}
\begin{tabular}{lc|cccc}
\toprule
Method & S (\%) & L1B & L3B & G2B & P3.5\\
\midrule
Zero-shot & 0 & 17.0 & 20.1 & 20.8 & 14.9 \\
LoRA & 100 & 43.2 & 47.8 & 47.9 & 45.4 \\
\hdashline
TA & 20 & 30.0 & 37.4 & 37.7 & 33.3 \\
TIES & 20 & 29.6 & 35.0 & 35.1 & 34.1 \\
DARE & 20 & 23.0 & 38.6 & 24.6 & 20.1 \\
DARE-TIES & 20 & 24.5 & 27.4 & 25.7 & 20.1 \\
\ourmethod(M=1) & 20 & 29.1 & 36.5 & 36.0 & 30.6 \\
\ourmethod(M=2) & 28 & 31.5 & 41.3 & 40.3 & 40.8 \\
\ourmethod(M=5) & 52 & 41.8 & 46.0 & 47.5 & 45.2 \\
\bottomrule
\end{tabular}
\end{center}
\end{table}

\noindent\textbf{Different low-rank-based adapter types:} 
In this experiment, we demonstrate how \ourmethod~can be extended to other low-rank-based adapters. In particular, we consider VeRA \cite{kopiczko2023vera} for its efficiency and competitive performance compared to LoRA. 

Similarly to the application of \ourmethod~to a set of LoRA parameters, we merge a set of VeRA parameters by learning a new set of parameters $\Lambda_b^\prime$ and $\Lambda_d^\prime$ with the latter shared across multiple tasks. 
Note that for simplicity, we discard the merging of the parameters $A$ and $B$ as they are initialized similarly across the tasks and kept frozen during fine-tuning. 

Table \ref{table:table3} presents the results. 
The best state-of-the-art approach is TIES in this case, while TA, which worked well on LoRA, achieves a significantly lower score here.
Even though TIES performs better than our method at 20\% storage efficiency, we remark once again that our approach allows us to achieve much higher performance at the minimal cost of 2.7\% additional storage.
\begin{table}[t]
\small
\setlength{\tabcolsep}{6pt}

\caption{\textbf{Average performance on 5 English applications using Llama-3.2-1B-Instruct VeRA-finetuned.} 
S represents the percentage of the parameters compared to storing 5 VeRAs.}
\label{table:table3}
\begin{center}
\begin{tabular}{lcc}
\toprule
Method & S (\%) & Avg\\
\midrule
Zero-shot & 0.0 & 17.2 \\
VeRA & 100.0 & 39.0 \\
\hdashline
TA & 20.0 & 0.3 \\
TIES & 20.0 & 27.8 \\
DARE & 20.0 & 28.6 \\
DARE-TIES & 20.0 & 27.7 \\
\ourmethod (M=1) & 20.0 & 26.9 \\
\ourmethod (M=2) & 20.7 & 27.5 \\
\ourmethod (M=3) & 21.4 & 29.9 \\
\ourmethod (M=4) & 22.1 & 35.8 \\
\ourmethod (M=5) & 22.7 & 36.4 \\
\bottomrule
\end{tabular}
\end{center}
\end{table}

\noindent\textbf{Evaluation across languages and merging setups:} We extend our setup to multiple languages in Table~\ref{table:table4}. 
Firstly, we merge 5 LoRA adapters in each language and report the average performance across 8 languages (\textit{application} block). 
Secondly, we apply merging across languages rather than applications, \ie, merging 8 LoRA adapters for each application (\textit{languages} block). 
Finally, we merge all 40 LoRA adapters, each of which corresponds to an application in a given language (\textit{task} block).
Detailed individual results are reported in Appendix \ref{sec:additional_results}. 

The results highlight that our method is in line with existing state-of-the-art merging methods for the highest storage efficiency levels, however, it enables large accuracy gains when small additional storage is available.

\begin{table}[t]
\small

\caption{\textbf{Average performance on 40 tasks using Llama-3.2-1B-Instruct (L1B) and Llama-3.2-3B-Instruct (L3B) LoRA-finetuned.} S represents the percentage of parameters compared to storing all 40 LoRAs.}
\label{table:table4}
\begin{center}
\hspace*{0.5em}
\begin{tabular}{lcccc}
\toprule
Method & S (\%) & L1B & L3B & Avg\\
\midrule
Zero-shot & 0.0 & 12.3 & 15.1 & 13.7\\
LoRA & 100.0 & 28.1 & 33.2 & 30.7 \\
\hdashline
\tikzmark[xshift=-4pt,yshift=1ex]{x1}TA & 20.0 & 19.3 & 24.1 & 21.7\\
TIES & 20.0 & 17.9 & 20.5 & 19.2 \\
DARE & 20.0 & 15.0 & 24.2 & 19.6 \\
DARE-TIES & 20.0 & 15.6 & 18.4 & 17.0 \\
\ourmethod(M=1) & 20.0 & 18.9 & 23.3 & 21.1 \\
\ourmethod(M=2) & 28.0 & 19.5 & 26.6 & 23.1 \\
\tikzmark[xshift=-4pt,yshift=-1ex]{y1}\ourmethod(M=5) & 52.0 & 26.9 & 32.4 & 29.6 \\
\hdashline
\tikzmark[xshift=-4pt,yshift=1ex]{x2}TA & 12.5 & 24.6 & 29.1 & 26.9 \\
TIES & 12.5 & 24.7 & 25.3 & 25.0\\
DARE & 12.5 & 15.8 & 19.0 & 17.4\\
DARE-TIES & 12.5 & 17.1 & 21.1 & 19.1 \\
\ourmethod(M=1) & 12.5 & 23.9 & 27.4 & 25.6\\
\ourmethod(M=3) & 22.5 & 24.7 & 29.7 & 27.2\\
\tikzmark[xshift=-4pt,yshift=-1ex]{y2}\ourmethod(M=8) & 47.5 & 26.6 & 32.1 & 29.4 \\
\hdashline
\tikzmark[xshift=-4pt,yshift=1ex]{x3}TA & 2.5 & 18.0 & 21.8 & 19.9\\
TIES & 2.5 & 17.4 & 18.7 & 18.0\\
DARE & 2.5 & 14.0 & 16.7 & 15.3\\
DARE-TIES & 2.5 & 14.2 & 16.6 & 15.4\\
\ourmethod(M=1) & 2.5 & 17.5 & 22.0 & 19.8\\
\tikzmark[xshift=-4pt,yshift=-1ex]{y3}\ourmethod(M=40) & 41.5 & 21.9 & 25.2 & 23.5\\
\bottomrule
\end{tabular}
\drawbrace[brace mirrored, thick]{x1}{y1}
\drawbrace[brace mirrored, thick]{x2}{y2}
\drawbraceshort[brace mirrored, thick]{x3}{y3}
\annote[left,rotate=90,violet]{brace-1}{applications}
\annote[left,rotate=90,teal]{brace-2}{languages}
\annote[left,rotate=90,purple]{brace-3}{tasks}
\end{center}
\end{table}

\noindent\underline{Merging Across Applications:} 
Our approach performs comparably to the best state-of-the-art method (TA), with only a 0.6\% drop in performance. However, by utilizing 8\% more storage, \ourmethod~achieves a 1.4\% performance gain over TA. Additionally, when a 48\% reduction in storage size is acceptable, the average performance drop compared to the upper bound individual LoRA adapters can be reduced to 1.1\%.

\noindent\underline{Merging Across Languages}: In this setting, TA achieves the best performance at 12.5\% storage efficiency, with an average performance drop of 3.8\% compared to individual LoRA adapters. \ourmethod~surpasses TA with just a 10\% increase in storage, and the performance gap continues to widen as storage size increases. At 47.5\% storage efficiency, \ourmethod~incurs only a 1.3\% drop compared to individual LoRA adapters.

\noindent\underline{Merging Across Tasks}: In this most challenging setup, TA and \ourmethod~exhibit similar performance at the most constrained storage efficiency of 2.5\%. However, as storage size increases to 41.5\%, our approach narrows the gap with individual LoRA adapters to as little as 7.2\%.

\subsection{Ablation Studies}
\label{sec:ablation_results}
We fine-tune Llama-3.2-1B-Instruct using the English data and perform several ablation studies when merging 5 LoRA adapters.

\noindent\textbf{Impact of LoRA rank:} First, we investigate whether the benefits of \ourmethod~remain the same across the LoRA rank $r \in \{8, 16, 32\}$. As shown in Table~\ref{table:table5}, \ourmethod~ begins to improve over the performance of the leading competitor method when the storage is increased by 8\%. Moreover, a 48\% reduction in storage size leads to competitive performance with storing all individual LoRA adapters.

\begin{table}[t]

\caption{\textbf{Impact of LoRA rank on average performance for 5 English applications using Llama-3.2-1B-Instruct LoRA-finetuned}. We report the percentage of the parameters compared to storing all 5 LoRA parameters (denoted by S) and the average score.}

\small

\label{table:table5}
\begin{center}
\resizebox{\columnwidth}{!}{%
\begin{tabular}{lcccc}
\toprule
Method & S (\%) & $r=8$ & $r=16$ & $r=32$\\
\midrule
Zero-shot & 0 & 17.0 & 17.0 & 17.0\\
LoRA & 100 & 39.2 & 41.7 & 43.2\\
\hdashline
TA & 20 & 28.4 & 28.9 & 30.0\\
TIES & 20 & 28.1 & 29.1 & 29.6\\
DARE & 20 & 20.7 & 21.8 & 23.0\\
DARE-TIES & 20 & 22.4 & 23.8 & 24.5\\
\ourmethod (M=1) & 20 & 27.1 & 28.3 & 29.1\\
\ourmethod (M=2) & 28 & 30.0 & 31.0 & 31.5\\
\ourmethod (M=5) & 52 & 38.5 & 40.0 & 41.8\\
\bottomrule
\end{tabular}
}
\end{center}
\end{table} 

\begin{figure}[t]
\centering
    \includegraphics[width=\linewidth]{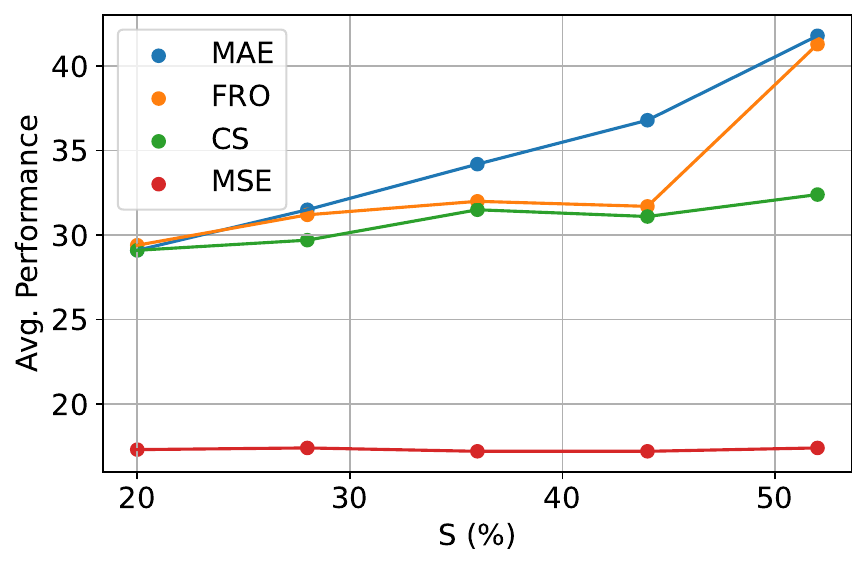}
  \caption{\textbf{Impact of distance function used during training.} We report average performance on 5 English applications using Llama-3.2-1B-Instruct LoRA-finetuned.}
  \label{fig:loss_functions}
\end{figure}

\noindent\textbf{Impact of the distance function:} Next, we analyze the choice of distance function $f$ used for calculating the loss during training. In particular, we compare mean absolute error (MAE) with three alternative loss functions based on cosine similarity (CS), Frobenius norm (FRO), and mean squared error (MSE). Fig.~\ref{fig:loss_functions} shows that MAE and FRO lead to better performance than CS and MSE. This result is not surprising as the optimization is done over the sparse LoRA parameters, which can be better learned using sparsity-inducing penalties \cite{bach2012optimization}.

\begin{figure}[b]
\centering
    \includegraphics[width=\linewidth]{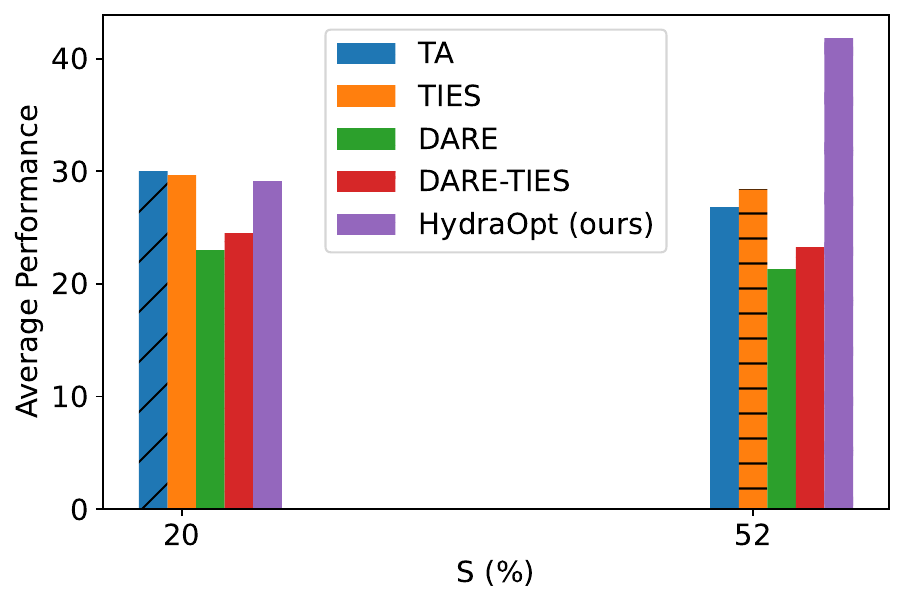}
  \caption{\textbf{Impact of storage efficiency level.} We report average performance on 5 English applications using Llama-3.2-1B-Instruct LoRA-finetuned.}
  \label{fig:storage_efficiency}
\end{figure}
\noindent\textbf{Analysis of existing merging methods at higher storage sizes:} We apply existing merging methods only on the $A$ parameters for comparison with our method at a reduced storage efficiency level. The existing merging methods lead to a performance drop as shown in Fig.~\ref{fig:storage_efficiency}, which can be explained by the lack of adaptation on the $B$ parameters after obtaining the new shared $A$ parameters. \ourmethod, on the other hand, iteratively adapts the $B^\prime$ parameters to the shared $A^\prime$ parameter to approximate the original model updates $\Delta W$.

\section{Conclusion}
On-device applications of LLMs often leverage parameter-efficient fine-tuning methods, such as low-rank-based adapters, for downstream tasks. However, the need to deploy a separate adapter for each task results in substantial storage overhead, a critical challenge for resource-constrained environments such as mobile devices. While model merging techniques offer a potential solution by reducing the storage size, they often come at the cost of significant performance degradation on downstream tasks, making them impractical for real-world deployment.

In this work, we introduce \ourmethod, a new model merging technique that effectively addresses the trade-off. \ourmethod~achieves competitive performance (0.2-1.8\% drop) compared to storing all LoRA adapters, while significantly reducing parameter size (48\% reduction). Furthermore, it consistently outperforms the performance of existing model merging methods at slightly worse efficiency levels. \ourmethod~thus enables efficient storage utilization without compromising task-specific performance for deploying LLMs on-device.

\section*{Limitations}
Despite the encouraging results obtained using \ourmethod, there are certain limitations in our current study that are worth acknowledging. For instance, this paper considers the generic case of data-free model merging where there is no assumption on the presence of data. Therefore, its performance is upper bounded by LoRA parameters. An interesting research direction may be considering data-driven merging scenarios, for which we expect similar gains in terms of efficiency and performance. Moreover, we tested \ourmethod~with low-rank-based adapters such as LoRA and VeRA due to their popularity and efficiency-performance trade-offs. Exploring its applications to other types of adapters would be a valuable direction for future work. Finally, we believe that our proposed method can be used in more general cross-modal and multi-modal tasks, which are getting more and more attention in the literature.

\section*{Potential Risks}
This work investigates the efficiency-performance trade-off of serving LLMs for text generation applications on mass accessible devices. Even though our work can reduce the storage costs of LLMs, it does not change the inference complexity and its impact on the environment. Moreover, we did not evaluate how our method impacts LLMs in terms of fairness across different population subgroups, which needs to be verified using additional safeguarding tools.

\bibliography{custom}

\clearpage 

\appendix
\begin{table*}[!htb]
\setlength{\tabcolsep}{3pt}
\caption{\textbf{Dataset statistics.} Information about the datasets.}
\label{table:datasets}
\begin{center}
\resizebox{\textwidth}{!}{%
\begin{tabular}{cccrrr}
\toprule
Task & Dataset & Language & \# Training Samples & \# Validation Samples & \# Test Samples \\
\midrule
\multirow{8}{*}{Grammar Error Correction} & \href{https://github.com/aopolin-lv/ECSpell/tree/main/Data/domains_data}{ECSpell} & Chinese & 6,680 & 750 & 750\\
& \href{https://www.cl.cam.ac.uk/research/nl/bea2019st/data/wi+locness_v2.1.bea19.tar.gz}{Cambridge English Write \& Improve (W\&I)} & English & 23,523 & 2,526 & 2,639 \\
& \href{https://www.merlin-platform.eu}{Merlin} & Italian & 572 & 79 & 81\\
\cline{2-6}
& \multirow{5}{*}{\href{https://github-typo-corpus.s3.amazonaws.com/data/github-typo-corpus.v1.0.0.jsonl.gz}{GitHub Typo Corpus}} & French & 616 & 240 & 227\\
& & German & 412 & 119 & 132\\
& & Japanese & 1,043 & 325 & 321\\
& & Korean & 255 & 75 & 93\\
& & Spanish & 348 & 137 & 116\\
\cline{1-6}
Smart Reply & \href{https://huggingface.co/datasets/google/Synthetic-Persona-Chat}{Persona-Chat Synthetic} & English & 225,061 & 25,847 & 24,479 \\
Text Summarization & \href{https://huggingface.co/datasets/Samsung/samsum}{SAMSum} & English & 14,732 & 818 & 819 \\
Tone Adjustment & \href{https://huggingface.co/datasets/facebook/content_rephrasing}{Sound Natural} & All & 2,245 & 321 & 642 \\
Question Answering & \href{https://huggingface.co/datasets/rajpurkar/squad}{SQuAD} & English & 65,699 & 21,900 & 10,570 \\
\bottomrule
\end{tabular}
}
\end{center}
\end{table*}

\section{Appendix}
\label{sec:appendix}
We provide information about the datasets in Sec.~\ref{sec:datasets},
report analyses about the similarity between LoRA parameters in Sec.~\ref{sec:lora_similarities}, and give more detailed results in Sec.~\ref{sec:additional_results}.

\subsection{Implementation Details}
\label{sec:datasets}
Table~\ref{table:datasets} provides a summary of the datasets used in our experiments and additional information about our implementation. Table~\ref{tab:prompts} describes the prompts that we have used. For tone adjustment, we consider four tones, namely professional, casual, witty and (neutral) paraphrasing, which are combined to create a larger dataset. We utilize NVIDIA A40 for our experiments. Adapters are applied to the query, key, value, and output matrices for Llama-3.2-3B-Instruct, Llama-3.2-1B-Instruct, and Phi-3.5-mini-instruct, and to the query and value for Gemma-2-2B-it. Uniform merging coefficients are used for TA and  DARE, whereas unary coefficients are used for TIES and  DARE-TIES. We provide an implementation of our method in Figures \ref{fig:hydraopt_module}, \ref{fig:hydraopt_loss}, \ref{fig:hydraopt_merging}. 

\begin{figure*}
\begin{lstlisting}[language=Python]
class HydraOpt(nn.Module):
    def __init__(
        self,
        lora_parameters,
        lora_names,
        M,
        T,
    ) -> None:
        super().__init__()

        self.K = len(lora_names)
        self.M = M
        self.T = T

        # trainable parameters 
        self.A_prime = nn.ModuleDict({})
        self.B_primes = nn.ModuleDict({})                
        self.C_primes = None if self.K == self.M else nn.ParameterDict({}) 
        
        keys = lora_parameters[lora_names[0]].keys()
        for i in range(M):
            peft_model_id = lora_names[i]
            for key in keys:
                # '.' can't be used in module_dict
                key_ = key.replace(".", "_")

                # initialize the shared A_prime  
                if i == 0 and "lora_A" in key:
                    r, in_features = lora_parameters[peft_model_id][key].shape
                    self.A_prime[key_] = nn.Linear(in_features, r, bias=False)
                    nn.init.kaiming_uniform_(
                            self.A_prime[key_].weight, a=math.sqrt(5)
                        )

                # initialize the B_primes 
                if "lora_B" in key:
                    out_features, r = lora_parameters[peft_model_id][key].shape
                    self.B_primes[f"{key_}_{i}"] = nn.Linear(
                        r, out_features, bias=False
                    )
                    nn.init.zeros_(self.B_primes[f"{key_}_{i}"].weight)

        # define coefficients if needed
        if self.K != self.M:
            for i in range(self.K):
                for key in keys:
                    key_ = key.replace(".", "_")
                    
                    if "lora_B" in key:
                        self.C_primes[f"{key_}_{i}"] = nn.Parameter(
                            nn.functional.softmax(
                                torch.randn(M) / T, dim=-1
                                )
                            )

    def forward(self):
        return self.A_prime, self.B_primes, self.C
        
\end{lstlisting}
\caption{\textbf{Implementation of HydraOpt module.}} 
\label{fig:hydraopt_module}
\end{figure*}

\begin{figure*}
\begin{lstlisting}[language=Python]
def hydra_loss(
    A_prime, B_primes, C_primes, M, lora_parameters, lora_names, T
):

    keys = list(lora_parameters[lora_names[0]].keys()) 
    K = len(lora_names)
    J = len(keys)
    
    for i, peft_model_id in enumerate(lora_names):
        for j in range(J):            
            if j % 2 == 0:
                A_key = keys[j]
                B_key = keys[j + 1]
                A_key_ = A_key.replace(".", "_")
                B_key_ = B_key.replace(".", "_")

                if "lora" not in A_key or "lora" not in B_key:
                    continue

                # calculate the original model update
                delta_W = lora_parameters[peft_model_id][B_key] @ lora_parameters[peft_model_id][A_key]

                if K == M:
                    delta_W_prime = (
                        B_primes[f"{B_key_}_{i}"].weight @ A_prime[A_key_].weight
                    )
                else: 
                    delta_W_prime = torch.stack(
                        [
                            nn.functional.softmax(
                                C_primes[f"{B_key_}_{i}"] * T, dim=-1
                                )[l]
                            * B_primes[f"{B_key_}_{l}"].weight
                            @ A_prime[A_key_].weight
                            for l in range(M)
                        ]
                    ).sum(dim=0)
                if i == 0 and j == 0:
                    total_loss = nn.functional.l1_loss(delta_W, delta_W_prime)
                else:
                    total_loss += nn.functional.l1_loss(delta_W, delta_W_prime)
    
    return total_loss / (K * J)
\end{lstlisting}
\caption{\textbf{Implementation of HydraOpt loss function.}} 
\label{fig:hydraopt_loss}
\end{figure*}

\begin{figure*}
\begin{lstlisting}[language=Python]
def hydraopt_merging(lora_parameters, lora_names, M, lr=0.01, epochs=1000, T=5.0):
    
    hydraopt = HydraOpt(lora_parameters, lora_names, M=M, T=T)
    hydraopt = hydraopt.to("cuda:0")
    
    keys = list(lora_parameters[lora_names[0]].keys())
    
    # move each pair of LoRA parameters to GPU
    for i, peft_model_id in enumerate(lora_names):
        for j in range(len(keys)):
            if j % 2 == 0:
                A_key = keys[j]
                B_key = keys[j + 1]
                if "lora" not in A_key or "lora" not in B_key:
                    continue
                else:
                    lora_parameters[peft_model_id][B_key] = lora_parameters[peft_model_id][B_key].to("cuda:0")
                    lora_parameters[peft_model_id][A_key] = lora_parameters[peft_model_id][A_key].to("cuda:0")

     
    optimizer = torch.optim.AdamW(hydraopt.parameters(), lr=lr)
    for epoch in range(epochs):
        optimizer.zero_grad()

        # get the current parameters
        A_prime, B_primes, C_primes = hydraopt()

        # calculate the loss based on the distance to the original LoRA updates 
        loss = hydra_loss(
            A_prime,
            B_primes,
            C_primes,
            M,
            lora_parameters,
            lora_names   
        )

        # backward pass
        loss.backward()

        # update weights
        optimizer.step()
    
    # obtain the mapping between the lora_Bs and B_primes parameters
    B_mapping = {}
    for task_index in range(len(lora_names)):
        B_mapping[task_index] = {}
        for key in keys:
            key_ = key.replace(".", "_")
            if "lora_B" in key:    
                if hydraopt.K == hydraopt.M:
                    selected_B_prime = task_index
                else:
                    selected_B_prime = nn.functional.softmax(
                        hydraopt.C_primes[f"{key_}_{task_index}"] / T, dim=-1
                        ).argmax()
                B_mapping[task_index][key] = selected_B_prime

    return A_prime, B_primes, B_mapping
\end{lstlisting}
\caption{\textbf{Implementation of HydraOpt merging.}} 
\label{fig:hydraopt_merging}
\end{figure*}

\begin{CJK}{UTF8}{mj}
\begin{table*}[h]
\begin{center}
\resizebox{0.98\textwidth}{!}{%
\begin{tabular}{lcc}
\toprule
Problem Type & Language & Prompt \\
\midrule
\multirow{8}{*}{Grammar Error Correction} & English & Remove all grammatical errors from this text: \\
& Spanish & Quita todos los errores gramaticales de este texto: \\
& French & Supprimez tous les erreurs grammaticales de ce texte: \\
& German & Verbessere alle grammatischen Fehler in diesem Text: \\
& Italian & Rimuovi tutti gli errori grammaticali da questo testo: \\
& Chinese & 删除该文本中的所有语法错误: \\
& Korean & 주어진 사용자의 입력에 오타나 문법 오류가 있으면 고친다: \\
& Japanese & このテキストからすべての文法エラーを削除する: \\
\midrule
\multirow{8}{*}{Smart Reply} & English & Suggest a reply for the following text: \\
& Spanish & Sugiera una respuesta para el texto siguiente: \\
& French & Propose une réponse pour le texte suivant: \\
& German & Schlagen Sie eine Antwort für den folgenden Text vor: \\
& Italian & Suggerisci una risposta per il seguente testo: \\
& Chinese & 建议对以下文本进行回复: \\
& Korean & 다음 텍스트에 대한 답변을 제안하시오: \\
& Japanese & 次のテキストに対する返信を提案します: \\
\midrule
\multirow{8}{*}{Text Summarization} & English & Summarize the following text: \\
& Spanish & Resume el siguiente texto: \\
& French & Résume le texte suivant: \\
& German & Zusammenfassen Sie den folgenden Text: \\
& Italian & Riassumi il seguente testo: \\
& Chinese & 总结一下下面的文字: \\
& Korean & 다음 텍스트를 요약하시오: \\
& Japanese & 次の文章を要約します: \\
\midrule
\multirow{8}{*}{Tone Adj. (Professional)} & English & Changes a given user's input sentence or text to the Professional style: \\
& Spanish & Cambia la oración o el texto introducido por un usuario al estilo Profesional: \\
& French & Transforme la phrase ou le texte saisi par un utilisateur en style Professionnel: \\
& German & ändert die Eingabe eines bestimmten Benutzers in einen Professionellen Stil: \\
& Italian & Cambia la frase o il testo immesso da un utente in stile Professionale: \\
& Chinese & 将给定用户的输入句子或文本更改为专业风格: \\
& Korean & 주어진 사용자의 입력을 전문적인 문체로 변경한다: \\
& Japanese & 指定されたユーザーの入力文またはテキストを プロフェッショナル スタイルに変更する: \\
\midrule
\multirow{8}{*}{Tone Adj. (Casual)} & English & Changes a given user's input sentence or text to the Casual style: \\
& Spanish & Cambia la oración o el texto introducido por un usuario al estilo Informal: \\
& French & Transforme la phrase ou le texte saisi par un utilisateur en style Informel: \\
& German & ändert die Eingabe eines bestimmten Benutzers in einen Freundlichen Stil: \\
& Italian & Cambia la frase o il testo immesso da un utente in stile Informal: \\
& Chinese & 将给定用户的输入句子或文本更改为日常风格: \\
& Korean & 주어진 사용자의 입력을 평범한 문체로 변경한다: \\
& Japanese & 指定されたユーザーの入力文またはテキストを カジュアル スタイルに変更する: \\
\midrule
\multirow{8}{*}{Tone Adj. (Witty)} & English & Changes a given user's input sentence or text to the Witty style: \\
& Spanish & Cambia la oración o el texto introducido por un usuario al estilo Ingenioso: \\
& French & Transforme la phrase ou le texte saisi par un utilisateur en style Spirituel: \\
& German & ändert die Eingabe eines bestimmten Benutzers in einen Witziger Stil: \\
& Italian & Cambia la frase o il testo immesso da un utente in stile Spiritoso: \\
& Chinese & 将给定用户的输入句子或文本更改为机智风格: \\
& Korean & 주어진 사용자의 입력을 재치있는 문체로 변경한다: \\
& Japanese & 指定されたユーザーの入力文またはテキストを ウィットに富んだ スタイルに変更する: \\
\midrule
\multirow{8}{*}{Tone Adj. (Paraphrase)} & English & Paraphrase the following text: \\
& Spanish & Parafrasea el siguiente texto: \\
& French & Paraphraser le texte suivant: \\
& German & Fassen Sie den folgenden Text zusammen: \\
& Italian & Parafrasare il testo seguente: \\
& Chinese & 解释以下文字: \\
& Korean & 다음 텍스트를 의역하세요: \\
& Japanese & 次のテキストを言い換えてください: \\
\midrule
\multirow{8}{*}{Question Answering} & English & Answer the following question: \\
& Spanish & Responde a la siguiente pregunta: \\
& French & Réponds à la question suivante: \\
& German & Beantworten Sie die folgende Frage: \\
& Italian & Rispondi alla seguente domanda: \\
& Chinese & 回答以下问题: \\
& Korean & 다음 질문에 답하시오: \\
& Japanese & 次の質問に答えましょう: \\
\bottomrule
\end{tabular}
}
\end{center}
\caption{\textbf{Prompts for each application and language.} Details about the prompts we have used.}
\label{tab:prompts}
\end{table*}
\end{CJK}

\subsection{Similarities Between LoRA Parameters}
\label{sec:lora_similarities}

\begin{figure}[!htb]
\centering
    \includegraphics[width=\linewidth]{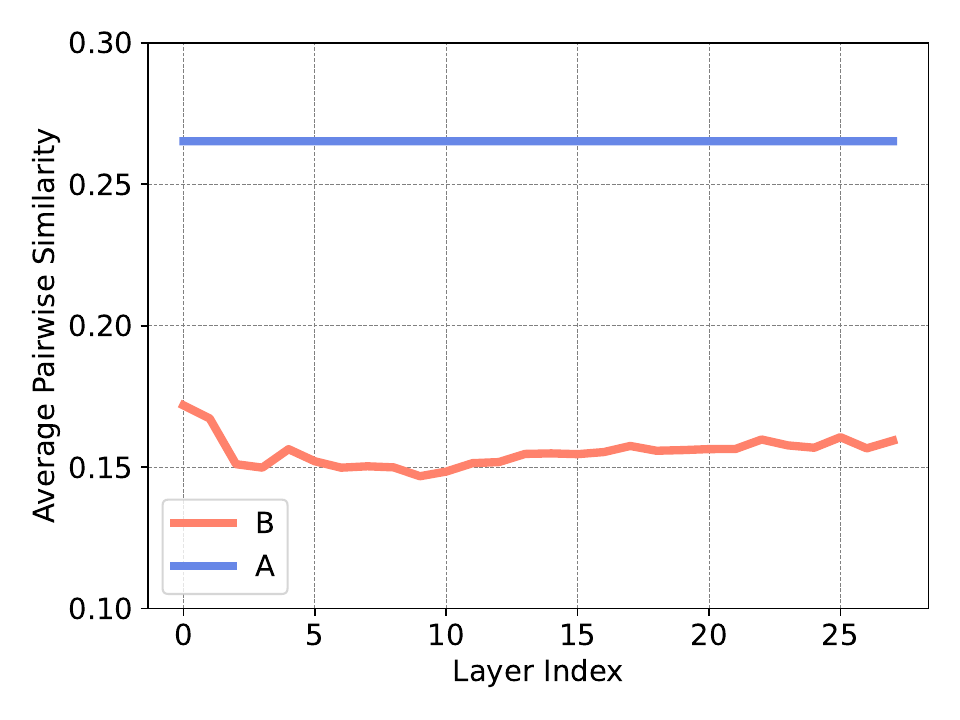}
  \caption{\textbf{Similarity between A and B matrices of LoRAs} measured using Canonical Correlation Analysis (CCA) goodness of fit as conducted by \citet{zhu2024asymmetry} on query matrices of Llama-3.2-3B-Instruct fine-tuned on English data.}  
  \label{fig:similarities_avg_cca}
\end{figure}

\begin{figure*}[!htb]
    \centering
    \includegraphics[width=\linewidth]{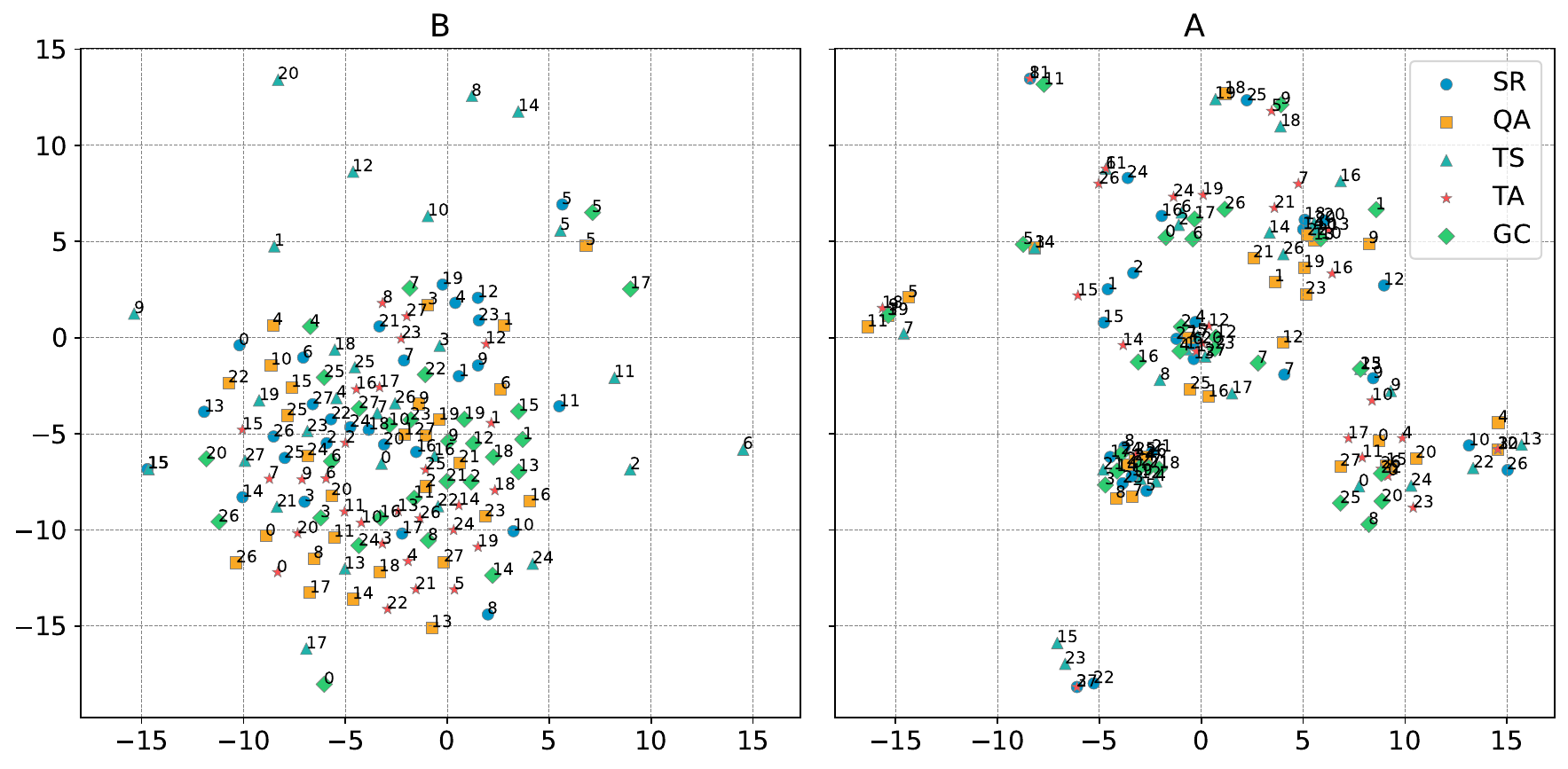}
    \caption{\textbf{t-SNE plots using LoRA parameters} for query matrices of Llama-3.2-3B-Instruct fine-tuned across 5 applications in English. Numbers indicate which layer the parameter comes from. Shapes/colors indicate the application the model is fine-tuned for. The differences in the LoRA parameters for different tasks arise mainly from the $B$ matrices.}
    \label{fig:tsne_results}
\end{figure*}

We perform Canonical Correlation Analysis (CCA) goodness of fit following \citet{zhu2024asymmetry} using LoRA matrices $A$ and $B$ obtained by fine-tuning Llama-3.2-3B-Instruct on English data. Fig.~\ref{fig:similarities_avg_cca} indicates that $A$ parameters result in higher similarity scores than $B$ parameters. Similarly, Fig.~\ref{fig:tsne_results} illustrates a similar trend using t-SNE plots where $A$ parameters tend to converge to similar values.

\subsection{Additional Results}
\label{sec:additional_results}

\begin{figure}[!htb]
    \includegraphics[width=\linewidth]{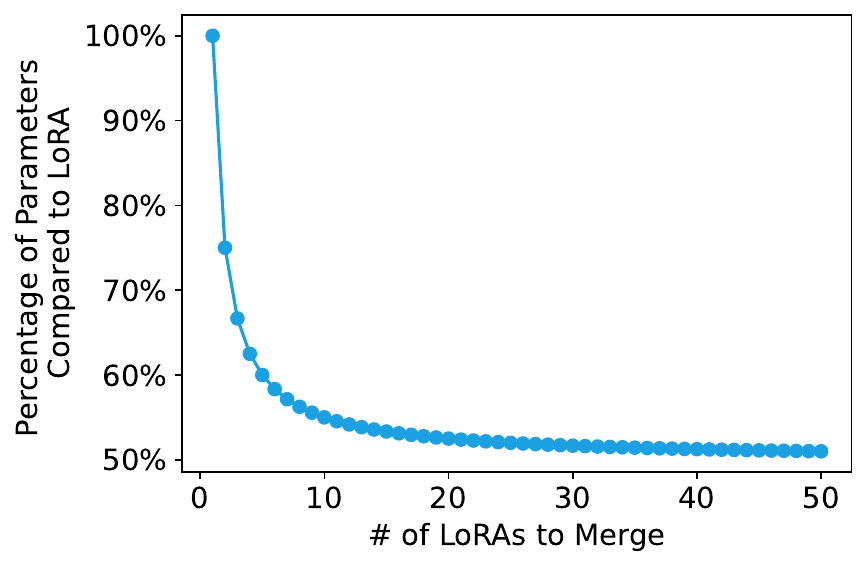}
  \caption{\textbf{The reduction in parameter size when using \ourmethod(M=K)}, assuming that the parameters $A$ and $B$ are the same size.}
  \label{fig:parameter_size_reduction}
\end{figure}

Fig.~\ref{fig:parameter_size_reduction} illustrates how the number of parameters to store changes using \ourmethod. We also present detailed evaluations for Llama-3.2-3B-Instruct when merging across applications (Tables~\ref{table:results_llama_3b_applications_detailed_1}-\ref{table:results_llama_3b_applications_detailed_2}), languages (Tables~\ref{table:results_llama_3b_languages_detailed_1}-\ref{table:results_llama_3b_languages_detailed_2}) and tasks (Table~\ref{table:results_llama_3b_tasks_detailed}). Moreover, the average scores are reported in Table \ref{table:results_llama_3b_detailed_avg}. Similarly, we present the detailed results under different ranks for Llama-3.2-3B-Instruct in Table \ref{table:merging_ranks}, and the results for Gemma-2-2B-it LoRA-finetuned (Table \ref{table:merging_gemma_2b}) and for Phi-3.5-mini-instruct LoRA-finetuned (Table \ref{table:merging_phi_3_5b}).Finally, Table \ref{table:runtime_memory} compares the methods in terms of runtime and GPU memory, which show that HydraOpt introduces additional runtime; however, the merging operation is still reasonably fast with relatively small GPU memory overhead.

\begin{table}[h!]
\caption{\textbf{Performance when merging across 5 applications using Llama-3.2-3B-Instruct for En (English), De (German), Es (Spanish), and Fr (French)}. Results are reported per each language (L) and application separately.}
\label{table:results_llama_3b_applications_detailed_1}
\begin{center}
\resizebox{.5\textwidth}{!}{%
\begin{tabular}{lcr|rrrrr|r}
\toprule
Method & L & S (\%) & GC & SR & TS & TA & QA & Avg\\
\midrule
Zero-shot & \multirow{11}{*}{En} & 0 & 14.0 & 5.4 & 26.9 & 30.1 & 24.2 & 20.1 \\
LoRA & & 100 & 39.2 & 24.5 & 41.4 & 59.1 &  74.7 & 47.8 \\
TA & & 20 & 27.0 & 13.6 & 34.1 & 53.7 & 58.8 & 37.4 \\
TIES & & 20 & 26.2 & 14.9 & 32.4 & 53.3 &  48.2 & 35.0 \\
DARE & & 20 & 27.4 & 14.9 & 34.5 & 54.2 & 61.9 & 38.6 \\
DARE TIES & & 20 & 21.6 & 8.3 & 29.2 & 45.2 & 33.0 & 27.4 \\
\ourmethod(M=1) & & 20 & 26.6 & 10.2 & 33.9 & 52.2 & 59.6 & 36.5 \\
\ourmethod(M=2) & & 30 & 27.4 & 22.9 & 34.6 & 54.8 & 67.0 & 41.3 \\
\ourmethod(M=3) & & 40 & 28.9 & 24.6 & 36.0 & 58.8 &  73.2 & 44.3 \\
\ourmethod(M=4) & & 50 & 36.0 & 24.7 & 39.6 & 57.6 & 72.3 & 46.0 \\
\ourmethod(M=5) & & 60 & 36.5 & 24.7 & 40.9 & 58.8 & 73.3 & 46.8 \\
\hdashline
Zero-shot & \multirow{11}{*}{De} & 0 &  9.4 & 2.8 & 17.7 & 20.3 & 10.9 & 12.2 \\
LoRA & & 100 & 41.2 & 13.8 & 32.4 & 44.8 &  47.2 & 35.9 \\
TA & & 20 & 22.9 & 6.7 & 23.6 & 41.2 &  22.6 & 23.4 \\
TIES & & 20 & 19.5 & 6.8 & 22.9 & 40.9 &  18.7 & 21.8 \\
DARE & & 20 & 26.4 & 6.6 & 20.5 & 42.7 & 29.4 & 25.1 \\
DARE TIES & & 20 & 12.1 & 4.1 & 20.9 & 30.0 & 15.6 & 16.5 \\
\ourmethod(M=1) & & 20 & 23.2 & 5.6 & 25.2 & 37.1 & 22.8 & 22.8 \\
\ourmethod(M=2) & & 30 & 25.8 & 9.6 & 26.2 & 43.4 & 34.8 & 28.0 \\
\ourmethod(M=3) & & 40 & 24.4 & 12.4 & 27.2 & 43.4 & 43.1 & 30.1 \\
\ourmethod(M=4) & & 50 & 24.6 & 13.4 & 29.7 & 43.7 & 45.3 & 31.3 \\
\ourmethod(M=5) & & 60 & 31.1 & 13.8 & 31.7 & 44.8 & 46.4 & 33.5 \\
\hdashline
Zero-shot & \multirow{11}{*}{Es} & 0 & 7.7 & 2.8 & 22.4 & 33.6 & 16.6 & 16.6 \\
LoRA & & 100 & 34.3 & 15.8 & 34.9 & 46.3 &  53.8 & 37.0 \\
TA & & 20 & 22.6 & 7.9 & 28.4 & 44.2 &  29.9 & 26.6 \\
TIES & & 20 & 16.9 & 8.7 & 27.0 & 44.3 &  23.7 & 24.1 \\
DARE & & 20 & 23.3 & 7.9 & 28.5 & 45.2 & 30.0 & 27.0 \\
DARE TIES & & 20 & 8.4 & 4.8 & 24.1 & 37.9 & 21.2 & 19.3 \\
\ourmethod(M=1) & & 20 & 20.8 & 6.3 & 29.2 & 42.5 & 28.6 & 25.5 \\
\ourmethod(M=2) & & 30 & 24.7 & 11.7 & 30.3 & 44.2 &  35.2 & 29.2 \\
\ourmethod(M=3) & & 40 & 25.3 & 15.1 & 31.5 & 45.4 & 46.8 & 32.8 \\
\ourmethod(M=4) & & 50 & 22.9 & 13.6 & 32.8 & 40.8 &  52.4 & 32.5 \\
\ourmethod(M=5) & & 60 & 31.2 & 15.1 & 34.5 & 46.1 &  53.1 & 36.0 \\
\hdashline
Zero-shot & \multirow{11}{*}{Fr} & 0 & 10.2 & 3.6 & 20.8 & 28.5 &  11.3 & 14.9 \\
LoRA & & 100 & 30.2 & 15.0 & 34.0 & 47.8 &  42.8 & 34.0 \\
TA & & 20 & 20.8 & 7.6 & 28.1 & 46.0 & 29.0 & 26.3 \\
TIES & & 20 & 16.4 & 8.8 & 27.6 & 46.1 &  25.8 & 24.9 \\
DARE & & 20 & 19.9 & 8.3 & 25.9 & 47.1 & 29.2 & 26.1 \\
DARE TIES & & 20 & 13.1 & 5.5 & 23.5 & 36.9 &  14.5 & 18.7 \\
\ourmethod(M=1) & & 20 & 19.3 & 6.8 & 28.7 & 43.3 &  27.4 & 25.1 \\
\ourmethod(M=2) & & 30 & 20.4 & 11.3 & 30.2 & 45.9  & 32.7 & 28.1 \\
\ourmethod(M=3) & & 40 & 22.4 & 13.9 & 30.6 & 46.8 & 37.7 & 30.3 \\
\ourmethod(M=4) & & 50 & 24.9 & 14.1 & 31.6 & 29.9 & 41.3 & 28.4 \\
\ourmethod(M=5) & & 60 & 30.8 & 14.9 & 33.8 & 47.8  & 42.4 & 34.0 \\
\bottomrule
\end{tabular}
}
\end{center}
\end{table}
\begin{table}[h!]
\caption{\textbf{Performance when merging across 5 applications using Llama-3.2-3B-Instruct for It (Italian), Ja (Japanese), Ko (Korean), and Zh (Chinese)}. Results are reported per each language (L) and application separately.}
\label{table:results_llama_3b_applications_detailed_2}
\begin{center}
\resizebox{.5\textwidth}{!}{%
\begin{tabular}{lcr|rrrrr|r}
\toprule
Method & L & S (\%) & GC & SR & TS & TA & QA & Avg\\
\midrule
Zero-shot & \multirow{11}{*}{It} & 0 & 26.9 & 2.3 & 18.9 & 29.5  & 13.5 & 18.2 \\
LoRA & & 100 & 36.0 & 12.7 & 32.3 & 44.9 &  47.3 & 34.6 \\
TA & & 20 & 34.0 & 6.0 & 25.3 & 43.4 &  21.0 & 26.0 \\
TIES & & 20 & 31.1 & 6.7 & 24.7 & 43.4 &  18.4 & 24.9 \\
DARE & & 20 & 29.7 & 6.0 & 24.9 & 44.5 &  23.8 & 25.8 \\
DARE TIES & & 20 & 27.6 & 3.7 & 20.1 & 36.6 & 15.9 & 20.8 \\
\ourmethod(M=1) & & 20 & 33.3 & 5.0 & 26.1 & 41.1 &  19.2 & 24.9 \\
\ourmethod(M=2) & & 30 & 33.2 & 8.5 & 26.8 & 43.0 & 31.9 & 28.7 \\
\ourmethod(M=3) & & 40 & 35.9 & 12.9 & 27.4 & 43.6 &  42.1 & 32.4 \\
\ourmethod(M=4) & & 50 & 33.5 & 12.2 & 29.8 & 35.0 & 46.2 & 31.3 \\
\ourmethod(M=5) & & 60 & 35.1 & 13.1 & 31.8 & 44.7 & 46.7 & 34.3 \\
\hdashline
Zero-shot & \multirow{11}{*}{Ja} & 0 & 4.1 & 4.7 & 19.5 & 35.6 & 12.6 & 15.3 \\
LoRA & & 100 & 23.3 & 11.9 & 29.2 & 40.1 &  31.1 & 27.1 \\
TA & & 20 & 10.9 & 6.0 & 25.0 & 40.2 &  20.7 & 20.6 \\
TIES & & 20 & 6.3 & 7.0 & 24.5 & 39.9  & 18.6 & 19.2 \\
DARE & & 20 & 9.4 & 5.8 & 25.2 & 40.0  & 22.0 & 20.5 \\
DARE TIES & & 20 & 7.1 & 5.7 & 22.0 & 38.0 &  14.0 & 17.3 \\
\ourmethod(M=1) & & 20 & 11.9 & 6.1 & 25.1 & 39.8 & 18.4 & 20.3 \\
\ourmethod(M=2) & & 30 & 13.3 & 8.2 & 25.2 & 40.2 &  22.7 & 21.9 \\
\ourmethod(M=3) & & 40 & 15.0 & 9.6 & 24.5 & 40.0 &  28.2 & 23.5 \\
\ourmethod(M=4) & & 50 & 18.1 & 11.2 & 26.4 & 37.7 &  26.3 & 24.0 \\
\ourmethod(M=5) & & 60 & 22.3 & 11.6 & 28.4 & 40.1 &  30.3 & 26.5 \\
\hdashline
Zero-shot & \multirow{11}{*}{Ko} & 0 & 7.7 & 0.9 & 9.4 & 25.7 & 11.5 & 11.0 \\
LoRA & & 100 & 16.6 & 5.8 & 15.6 & 31.6 &  24.8 & 18.9 \\
TA & & 20 & 6.9 & 2.1 & 13.0 & 31.2 &  11.8 & 13.0 \\
TIES & & 20 & 8.6 & 2.4 & 12.4 & 31.0 & 8.3 & 12.5 \\
DARE & & 20 & 7.1 & 2.2 & 13.2 & 31.2 & 9.9 & 12.7 \\
DARE TIES & & 20 & 7.9 & 1.7 & 10.7 & 28.0 & 12.3 & 12.1 \\
\ourmethod(M=1) & & 20 & 7.8 & 1.9 & 12.4 & 30.6 & 12.3 & 13.0 \\
\ourmethod(M=2) & & 30 & 9.0 & 3.0 & 12.4 & 31.2 &  15.7 & 14.2 \\
\ourmethod(M=3) & & 40 & 8.8 & 4.2 & 13.2 & 25.9 &  22.5 & 14.9 \\
\ourmethod(M=4) & & 50 & 9.4 & 5.1 & 13.2 & 29.7 &  22.2 & 15.9 \\
\ourmethod(M=5) & & 60 & 15.4 & 5.7 & 15.7 & 31.4 & 24.5 & 18.5 \\
\hdashline
Zero-shot & \multirow{11}{*}{Zh} & 0 & 3.7 & 1.6 & 20.2 & 24.8 & 12.1 & 12.5 \\
LoRA & & 100 & 48.6 & 7.9 & 27.6 & 37.3 &  30.1 & 30.3 \\
TA & & 20 & 10.6 & 3.7 & 24.1 & 37.8 &  19.9 & 19.2 \\
TIES & & 20 & 10.3 & 3.9 & 23.6 & 37.0 &  18.4 & 18.6 \\
DARE & & 20 & 6.3 & 3.7 & 23.4 & 37.9 & 19.3 & 18.1 \\
DARE TIES & & 20 & 6.2 & 2.4 & 21.0 & 30.2 &  14.6 & 14.9 \\
\ourmethod(M=1) & & 20 & 11.7 & 3.4 & 24.2 & 36.2 &  18.0 & 18.7 \\
\ourmethod(M=2) & & 30 & 17.0 & 4.9 & 24.9 & 38.0 & 23.2 & 21.6 \\
\ourmethod(M=3) & & 40 & 38.1 & 7.0 & 24.7 & 37.9  & 27.8 & 27.1 \\
\ourmethod(M=4) & & 50 & 44.0 & 6.4 & 24.9 & 35.7 &  28.3 & 27.8 \\
\ourmethod(M=5) & & 60 & 46.4 & 7.3 & 27.1 & 37.1 & 29.6 & 29.5 \\
\bottomrule
\end{tabular}
}
\end{center}
\end{table}
\begin{table}[h!]
\caption{\textbf{Performance when merging across 8 languages using Llama-3.2-3B-Instruct}. Results are reported per each language (L) and application separately.}
\label{table:results_llama_3b_languages_detailed_1}
\begin{center}
\resizebox{0.5\textwidth}{!}{%
\begin{tabular}{lcr|rrrrr|r}
\toprule
Method & L & S (\%) & GC & SR & TS & TA & QA & Avg\\
\midrule
Zero-shot & \multirow{14}{*}{En} & 0 & 14.0 & 5.4 & 26.9 & 30.1 & 24.2 & 20.1 \\
LoRA & & 100.0 & 39.2 & 24.5 & 41.4 & 59.1 & 74.7 & 47.8 \\
TA & & 12.5 & 27.6 & 15.3 & 35.8 & 49.6 & 75.5 & 40.8 \\
TIES & & 12.5 & 28.2 & 16.3 & 36.0 & 49.7 & 76.2 & 41.3 \\
DARE & & 12.5 & 17.2 & 7.6 & 29.4 & 35.5 & 34.5 & 24.9 \\
DARE TIES  & & 12.5 & 19.4 & 8.6 & 30.5 & 40.0 & 41.6 & 28.1 \\
\ourmethod(M=1) & & 12.5 & 26.1 & 14.1 & 34.4 & 48.1 & 73.4 & 39.2 \\
\ourmethod(M=2) & & 17.5 & 27.1 & 16.7 & 35.3 & 50.3 & 72.6 & 40.4 \\
\ourmethod(M=3) & & 22.5 & 29.5 & 17.9 & 36.9 & 51.6 & 71.2 & 41.4 \\
\ourmethod(M=4) & & 27.5 & 29.3 & 19.4 & 37.1 & 53.4 & 69.8 & 41.8 \\
\ourmethod(M=5) & & 32.5 & 36.2 & 21.2 & 41.2 & 57.0 & 63.0 & 43.7 \\
\ourmethod(M=6) & & 37.5 & 38.9 & 23.4 & 40.5 & 58.2 & 67.0 & 45.6 \\
\ourmethod(M=7) & & 42.5 & 39.1 & 24.1 & 41.0 & 57.9 & 72.4 & 46.9 \\
\ourmethod(M=8) & & 47.5 & 36.9 & 23.9 & 41.1 & 58.5 & 73.3 & 46.7 \\
\hdashline
Zero-shot & \multirow{14}{*}{De} & 0 & 9.4 & 2.8 & 17.7 & 20.3 &  10.9 & 12.2 \\
LoRA & & 100.0 & 41.2 & 13.8 & 32.4 & 44.8 & 47.2 & 35.9 \\
TA & & 12.5 & 30.5 & 7.7 & 31.5 & 41.2 & 46.2 & 31.4 \\
TIES & & 12.5 & 29.4 & 7.7 & 31.7 & 41.5 & 46.5 & 31.3 \\
DARE & & 12.5 & 12.4 & 4.2 & 21.9 & 28.5 & 17.3 & 16.9 \\
DARE TIES  & & 12.5 & 14.1 & 4.9 & 24.2 & 31.3 & 21.8 & 19.2 \\
\ourmethod(M=1)  & & 12.5 & 25.8 & 7.2 & 29.9 & 39.7 & 47.1 & 30.0 \\
\ourmethod(M=2) & & 17.5 & 26.8 & 7.8 & 30.6 & 41.6 & 46.9 & 30.7 \\
\ourmethod(M=3) & & 22.5 & 27.1 & 8.0 & 30.6 & 43.0 & 48.0 & 31.3 \\
\ourmethod(M=4) & & 27.5 & 25.7 & 8.2 & 30.6 & 42.4 & 45.9 & 30.5 \\
\ourmethod(M=5) & & 32.5 & 32.3 & 9.7 & 31.4 & 43.5 & 48.7 & 33.1 \\
\ourmethod(M=6) & & 37.5 & 26.9 & 8.6 & 31.2 & 43.9 & 45.7 & 31.3 \\
\ourmethod(M=7) & & 42.5 & 31.4 & 9.3 & 31.5 & 44.7 & 44.8 & 32.4 \\
\ourmethod(M=8) & & 47.5 & 31.0 & 12.1 & 31.6 & 44.5 & 46.3 & 33.1 \\
\hdashline
Zero-shot & \multirow{14}{*}{Es} & 0 & 7.7 & 2.8 & 22.4 & 33.6 &  16.6 & 16.6 \\
LoRA & & 100.0 & 34.3 & 15.8 & 34.9 & 46.3 & 53.8 & 37.0 \\
TA & & 12.5 & 32.6 & 9.4 & 34.6 & 34.0 & 48.5 & 31.9 \\
TIES & & 12.5 & 33.1 & 9.0 & 34.8 & 29.7 & 49.3 & 31.2 \\
DARE & & 12.5 & 9.8 & 4.9 & 26.8 & 37.8 & 24.9 & 20.8 \\
DARE TIES  & & 12.5 & 12.0 & 5.8 & 28.8 & 39.2 & 31.7 & 23.5 \\
\ourmethod(M=1)  & & 12.5 & 26.1 & 9.0 & 33.1 & 32.6 & 45.6 & 29.3 \\
\ourmethod(M=2) & & 17.5 & 28.6 & 9.4 & 33.5 & 43.6 & 48.0 & 32.6 \\
\ourmethod(M=3) & & 22.5 & 28.4 & 10.3 & 33.9 & 44.2 & 49.3 & 33.2 \\
\ourmethod(M=4) & & 27.5 & 26.6 & 11.4 & 34.6 & 45.7 & 53.1 & 34.3 \\
\ourmethod(M=5) & & 32.5 & 29.2 & 10.4 & 34.7 & 45.7 & 51.5 & 34.3 \\
\ourmethod(M=6) & & 37.5 & 28.6 & 12.0 & 34.1 & 45.1 & 53.3 & 34.6 \\
\ourmethod(M=7) & & 42.5 & 27.3 & 11.3 & 33.7 & 37.8 & 52.9 & 32.6 \\
\ourmethod(M=8) & & 47.5 & 31.7 & 13.5 & 34.5 & 46.1 & 52.5 & 35.7 \\
\hdashline
Zero-shot & \multirow{14}{*}{Fr} & 0 & 10.2 & 3.6 & 20.8 & 28.5 &  11.3 & 14.9 \\
LoRA & & 100.0 & 30.2 & 15.0 & 34.0 & 47.8 & 42.8 & 34.0 \\
TA & & 12.5 & 29.3 & 9.4 & 34.1 & 45.3 & 41.0 & 31.8 \\
TIES & & 12.5 & 30.4 & 9.6 & 34.4 & 45.0 & 42.2 & 32.3 \\
DARE & & 12.5 & 13.5 & 5.5 & 24.8 & 35.7 & 18.0 & 19.5 \\
DARE TIES  & & 12.5 & 14.3 & 6.4 & 26.5 & 37.6 & 26.3 & 22.2 \\
\ourmethod(M=1)  & & 12.5 & 24.2 & 9.2 & 32.6 & 44.0 & 38.5 & 29.7 \\
\ourmethod(M=2) & & 17.5 & 29.8 & 9.1 & 33.5 & 45.2 & 39.1 & 31.3 \\
\ourmethod(M=3) & & 22.5 & 26.7 & 9.7 & 33.8 & 45.9 & 40.8 & 31.4 \\
\ourmethod(M=4) & & 27.5 & 29.5 & 9.4 & 33.6 & 46.9 & 41.9 & 32.3 \\
\ourmethod(M=5) & & 32.5 & 31.6 & 9.6 & 33.7 & 47.4 & 40.2 & 32.5 \\
\ourmethod(M=6) & & 37.5 & 31.1 & 10.3 & 33.8 & 47.8 & 41.8 & 32.9 \\
\ourmethod(M=7) & & 42.5 & 30.3 & 10.2 & 33.0 & 46.8 & 39.0 & 31.9 \\
\ourmethod(M=8) & & 47.5 & 31.4 & 13.6 & 33.9 & 47.6 & 41.9 & 33.7 \\
\bottomrule
\end{tabular}
}
\end{center}
\end{table}
\begin{table}[h!]
\caption{\textbf{Performance when merging across 8 languages using Llama-3.2-3B-Instruct}. Results are reported per each language (L) and application separately.}
\label{table:results_llama_3b_languages_detailed_2}
\begin{center}
\resizebox{0.5\textwidth}{!}{%
\begin{tabular}{lcr|rrrrr|r}
\toprule
Method & L & S (\%) & GC & SR & TS & TA & QA & Avg\\
\midrule
Zero-shot & \multirow{14}{*}{It} & 0 & 26.9 & 2.3 & 18.9 & 29.5 & 13.5 & 18.2 \\
LoRA & & 100.0 & 36.0 & 12.7 & 32.3 & 44.9 & 47.3 & 34.6 \\
TA & & 12.5 & 34.0 & 6.3 & 32.4 & 42.8 & 45.3 & 32.2 \\
TIES & & 12.5 & 34.2 & 6.1 & 32.6 & 42.4 & 46.3 & 32.3 \\
DARE & & 12.5 & 28.6 & 3.6 & 21.6 & 35.9 & 18.9 & 21.7 \\
DARE TIES  & & 12.5 & 30.3 & 4.1 & 23.7 & 37.1 & 23.3 & 23.7 \\
\ourmethod(M=1)  & & 12.5 & 33.1 & 5.9 & 31.1 & 41.8 & 43.1 & 31.0 \\
\ourmethod(M=2) & & 17.5 & 33.8 & 6.6 & 30.9 & 42.1 & 45.0 & 31.7 \\
\ourmethod(M=3) & & 22.5 & 35.0 & 6.3 & 32.1 & 42.6 & 44.3 & 32.1 \\
\ourmethod(M=4) & & 27.5 & 35.3 & 7.4 & 31.0 & 43.1 & 45.5 & 32.4 \\
\ourmethod(M=5) & & 32.5 & 35.5 & 7.7 & 31.0 & 43.4 & 46.1 & 32.7 \\
\ourmethod(M=6) & & 37.5 & 34.7 & 9.7 & 31.8 & 43.8 & 45.5 & 33.1 \\
\ourmethod(M=7) & & 42.5 & 35.8 & 6.8 & 31.5 & 42.6 & 45.6 & 32.5 \\
\ourmethod(M=8) & & 47.5 & 35.0 & 11.4 & 31.7 & 44.5 & 46.3 & 33.8 \\
\hdashline
Zero-shot & \multirow{14}{*}{Ja} & 0 & 4.1 & 4.7 & 19.5 & 35.6 & 12.6 & 15.3 \\
LoRA & & 100.0 & 23.3 & 11.9 & 29.2 & 40.1 & 31.1 & 27.1 \\
TA & & 12.5 & 18.8 & 6.1 & 27.3 & 40.4 & 29.6 & 24.5 \\
TIES & & 12.5 & 21.5 & 5.9 & 27.5 & 39.7 & 30.4 & 25.0 \\
DARE & & 12.5 & 8.9 & 5.4 & 22.5 & 37.9 & 17.6 & 18.5 \\
DARE TIES  & & 12.5 & 11.1 & 5.9 & 23.6 & 38.5 & 22.3 & 20.3 \\
\ourmethod(M=1)  & & 12.5 & 15.0 & 6.1 & 26.3 & 40.2 & 28.6 & 23.2 \\
\ourmethod(M=2) & & 17.5 & 16.6 & 6.4 & 26.7 & 40.8 & 28.7 & 23.8 \\
\ourmethod(M=3) & & 22.5 & 18.3 & 6.9 & 26.4 & 40.4 & 28.7 & 24.1 \\
\ourmethod(M=4) & & 27.5 & 19.1 & 7.3 & 27.4 & 40.9 & 29.5 & 24.8 \\
\ourmethod(M=5) & & 32.5 & 20.8 & 8.0 & 28.4 & 40.8 & 29.1 & 25.4 \\
\ourmethod(M=6) & & 37.5 & 21.3 & 7.9 & 28.7 & 38.9 & 28.5 & 25.0 \\
\ourmethod(M=7) & & 42.5 & 22.2 & 5.0 & 28.3 & 37.5 & 29.9 & 24.6 \\
\ourmethod(M=8) & & 47.5 & 21.6 & 9.8 & 28.1 & 40.2 & 30.0 & 25.9 \\
\hdashline
Zero-shot & \multirow{14}{*}{Ko} & 0 & 7.7 & 9.4 & 25.7 & 4.2 & 11.5 & 11.0 \\
LoRA & & 100.0 & 16.6 & 5.8 & 15.6 & 31.6 & 24.8 & 18.9 \\
TA & & 12.5 & 12.8 & 2.0 & 14.0 & 31.3 & 19.7 & 16.0 \\
TIES & & 12.5 & 13.8 & 1.8 & 14.7 & 30.9 & 20.2 & 16.3 \\
DARE & & 12.5 & 12.8 & 1.6 & 11.3 & 28.0 & 17.4 & 14.2 \\
DARE TIES  & & 12.5 & 13.9 & 1.6 & 11.8 & 28.7 & 19.2 & 15.0 \\
\ourmethod(M=1)  & & 12.5 & 14.1 & 2.1 & 13.3 & 30.6 & 19.4 & 15.9 \\
\ourmethod(M=2) & & 17.5 & 13.8 & 2.4 & 13.7 & 31.0 & 20.8 & 16.3 \\
\ourmethod(M=3) & & 22.5 & 16.1 & 2.5 & 14.3 & 30.9 & 22.8 & 17.3 \\
\ourmethod(M=4) & & 27.5 & 13.4 & 2.9 & 14.5 & 31.8 & 21.4 & 16.8 \\
\ourmethod(M=5) & & 32.5 & 16.5 & 3.8 & 15.3 & 30.5 & 22.6 & 17.7 \\
\ourmethod(M=6) & & 37.5 & 11.2 & 3.1 & 15.0 & 31.1 & 25.3 & 17.1 \\
\ourmethod(M=7) & & 42.5 & 13.3 & 2.5 & 15.4 & 29.2 & 23.2 & 16.7 \\
\ourmethod(M=8) & & 47.5 & 15.2 & 4.5 & 15.2 & 31.3 & 24.5 & 18.2 \\
\hdashline
Zero-shot & \multirow{14}{*}{Zh} & 0 & 3.7 & 1.6 & 20.2 & 24.8 & 12.1 & 12.5 \\
LoRA & & 100.0 & 48.6 & 7.9 & 27.6 & 37.3 & 30.1 & 30.3 \\
TA & & 12.5 & 24.6 & 4.0 & 27.1 & 37.3 & 28.6 & 24.3 \\
TIES & & 12.5 & 28.4 & 4.0 & 27.4 & 37.4 & 29.2 & 25.3 \\
DARE & & 12.5 & 6.5 & 2.2 & 21.9 & 30.1 & 16.5 & 15.4 \\
DARE TIES  & & 12.5 & 7.8 & 2.6 & 23.0 & 31.4 & 20.3 & 17.0 \\
\ourmethod(M=1)  & & 12.5 & 12.6 & 3.6 & 26.1 & 36.4 & 26.5 & 21.0 \\
\ourmethod(M=2) & & 17.5 & 19.6 & 3.9 & 26.7 & 36.8 & 28.0 & 23.0 \\
\ourmethod(M=3) & & 22.5 & 38.3 & 4.4 & 26.6 & 38.2 &  28.2 & 27.2 \\
\ourmethod(M=4) & & 27.5 & 41.9 & 5.0 & 27.0 & 37.5 & 29.2 & 28.1 \\
\ourmethod(M=5) & & 32.5 & 43.1 & 5.2 & 27.1 & 37.6 & 27.7 & 28.1 \\
\ourmethod(M=6) & & 37.5 & 43.8 & 4.9 & 27.3 & 35.1 & 29.2 & 28.1 \\
\ourmethod(M=7) & & 42.5 & 43.4 & 4.7 & 24.3 & 36.1 & 26.2 & 26.9 \\
\ourmethod(M=8) & & 47.5 & 46.2 & 6.9 & 27.0 & 37.3 & 29.6 & 29.4 \\
\bottomrule
\end{tabular}
}
\end{center}
\end{table}
\begin{table*}[h!]
\caption{\textbf{Performance when merging across 40 tasks using Llama-3.2-3B-Instruct}. Results are reported per each language (L) and application separately.}

\label{table:results_llama_3b_tasks_detailed}
\small
\begin{center}
\resizebox{.7\textwidth}{!}{%
\begin{tabular}{lcr|rrrrr|r}
\toprule
Method & L & S (\%) & GC & SR & TS & TA & QA & Avg\\
\midrule
Zero-shot & \multirow{8}{*}{En} & 0 & 14.0 & 5.4 & 26.9 & 30.1 & 24.2 & 20.1 \\
LoRA & & 100.0 & 39.2 & 24.5 & 41.4 & 59.1 & 74.7 & 47.8 \\
TA & & 2.5 & 23.9 & 8.7 & 30.6 & 49.0 & 40.5 & 30.5 \\
TIES & & 2.5 & 23.3 & 8.7 & 30.3 & 48.2 & 35.9 & 29.3 \\
DARE & & 2.5 & 15.7 & 6.6 & 28.2 & 33.7 & 25.9 & 22.0 \\
DARE-TIES & & 2.5 & 15.6 & 6.6 & 28.2 & 33.7 & 25.8 & 22.0 \\
\ourmethod(M=1) & & 2.5 & 24.2 & 7.8 & 30.3 & 49.1 & 42.3 & 30.8 \\
\ourmethod(M=40) & & 41.5 & 28.9 & 22.1 & 35.6 & 57.0 & 71.3 & 43.0 \\
\hdashline
Zero-shot & \multirow{8}{*}{De} & 0 &  9.4 & 2.8 & 17.7 & 20.3 & 10.9 & 12.2 \\
LoRA & & 100.0 & 41.2 & 13.8 & 32.4 & 44.8 & 47.2 & 35.9 \\
TA & & 2.5 & 20.6 & 6.0 & 24.9 & 35.5 & 19.4 & 21.3 \\
TIES & & 2.5 & 17.5 & 6.0 & 24.2 & 34.9 & 18.1 & 20.2 \\
DARE & & 2.5 & 11.5 & 3.2 & 19.3 & 25.3 & 12.6 & 14.4 \\
DARE-TIES & & 2.5 & 11.8 & 3.3 & 19.1 & 25.4 & 12.5 & 14.4 \\
\ourmethod(M=1) & & 2.5 & 22.7 & 5.6 & 25.6 & 34.3 & 21.3 & 21.9 \\
\ourmethod(M=40) & & 41.5 & 20.3 & 7.2 & 27.7 & 42.4 & 31.2 & 25.8 \\
\hdashline
Zero-shot & \multirow{8}{*}{Es} & 0 & 7.7 & 2.8 & 22.4 & 33.6 &  16.6 & 16.6 \\
LoRA & & 100.0 & 34.3 & 15.8 & 34.9 & 46.3 & 53.8 & 37.0 \\
TA & & 2.5 & 19.2 & 6.8 & 28.9 & 41.5 & 25.4 & 24.4 \\
TIES & & 2.5 & 15.5 & 6.5 & 28.3 & 40.0 & 22.8 & 22.6 \\
DARE & & 2.5 & 7.9 & 3.7 & 24.0 & 36.6 & 19.7 & 18.4 \\
DARE-TIES & & 2.5 & 7.4 & 3.8 & 23.6 & 36.2 & 19.1 & 18.0 \\
\ourmethod(M=1) & & 2.5 & 20.3 & 6.2 & 29.2 & 42.3 & 27.6 & 25.1 \\
\ourmethod(M=40) & & 41.5 & 17.7 & 8.8 & 31.5 & 42.6 & 24.4 & 25.0 \\
\hdashline
Zero-shot & \multirow{8}{*}{Fr} & 0 & 10.2 & 3.6 & 20.8 & 28.5 & 11.3 & 14.9 \\
LoRA & & 100.0 & 30.2 & 15.0 & 34.0 & 47.8 & 42.8 & 34.0 \\
TA & & 2.5 & 17.5 & 7.2 & 27.5 & 43.2 & 23.3 & 23.8 \\
TIES & & 2.5 & 15.8 & 7.2 & 26.7 & 42.1 & 20.4 & 22.4 \\
DARE & & 2.5 & 12.4 & 4.5 & 22.1 & 33.5 & 12.1 & 16.9 \\
DARE-TIES & & 2.5 & 12.3 & 4.6 & 22.0 & 33.7 & 11.8 & 16.9 \\
\ourmethod(M=1) & & 2.5 & 19.2 & 6.8 & 28.9 & 42.1 & 24.6 & 24.3 \\
\ourmethod(M=40) & & 41.5 & 23.3 & 8.4 & 31.2 & 45.2 & 27.1 & 27.0 \\
\hdashline
Zero-shot & \multirow{8}{*}{It} & 0 & 26.9 & 2.3 & 18.9 & 29.5 & 13.5 & 18.2 \\
LoRA & & 100.0 & 36.0 & 12.7 & 32.3 & 44.9 & 47.3 & 34.6 \\
TA & & 2.5 & 33.0 & 4.9 & 26.7 & 41.7 & 19.7 & 25.2 \\
TIES & & 2.5 & 30.5 & 4.8 & 26.0 & 40.6 & 18.1 & 24.0 \\
DARE & & 2.5 & 26.7 & 2.7 & 19.7 & 34.6 & 15.2 & 19.8 \\
DARE-TIES & & 2.5 & 26.6 & 2.7 & 19.6 & 34.7 & 14.9 & 19.7 \\
\ourmethod(M=1) & & 2.5 & 33.5 & 5.1 & 27.2 & 41.2 & 19.1 & 25.2 \\
\ourmethod(M=40) & & 41.5 & 33.2 & 5.9 & 29.2 & 41.4 & 20.7 & 26.1 \\
\hdashline
Zero-shot & \multirow{8}{*}{Ja} & 0 & 4.1 & 4.7 & 19.5 & 35.6 &  12.6 & 15.3 \\
LoRA & & 100.0 & 23.3 & 11.9 & 29.2 & 40.1 & 31.1 & 27.1 \\
TA & & 2.5 & 9.8 & 6.2 & 22.5 & 39.3 & 18.9 & 19.3 \\
TIES & & 2.5 & 9.4 & 6.0 & 22.3 & 39.2 & 18.0 & 19.0 \\
DARE & & 2.5 & 6.3 & 4.9 & 20.7 & 37.2 & 13.0 & 16.4 \\
DARE-TIES & & 2.5 & 6.1 & 5.0 & 20.8 & 37.3 & 12.9 & 16.4 \\
\ourmethod(M=1) & & 2.5 & 11.6 & 6.2 & 23.3 & 39.3 & 17.7 & 19.6 \\
\ourmethod(M=40) & & 41.5 & 13.8 & 6.4 & 23.3 & 39.1 & 24.5 & 21.4 \\
\hdashline
Zero-shot & \multirow{8}{*}{Ko} & 0 & 7.7 & 0.9 & 9.4 & 25.7 &  11.5 & 11.0 \\
LoRA & & 100.0 & 16.6 & 5.8 & 15.6 & 31.6 & 24.8 & 18.9 \\
TA & & 2.5 & 9.0 & 1.6 & 10.6 & 30.7 & 13.1 & 13.0 \\
TIES & & 2.5 & 8.1 & 1.8 & 10.6 & 30.6 & 12.3 & 12.7 \\
DARE & & 2.5 & 9.1 & 1.3 & 10.1 & 27.4 & 12.1 & 12.0 \\
DARE-TIES & & 2.5 & 8.9 & 1.4 & 10.0 & 27.5 & 11.9 & 11.9 \\
\ourmethod(M=1) & & 2.5 & 8.4 & 1.6 & 11.2 & 30.3 & 13.5 & 13.0 \\
\ourmethod(M=40) & & 41.5 & 13.8 & 1.8 & 10.7 & 30.8 & 15.5 & 14.5 \\
\hdashline
Zero-shot & \multirow{8}{*}{Zh} & 0 & 3.7 & 1.6 & 20.2 & 24.8 & 12.1 & 12.5 \\
LoRA & & 100.0 & 48.6 & 7.9 & 27.6 & 37.3 & 30.1 & 30.3 \\
TA & & 2.5 & 7.9 & 3.1 & 22.6 & 35.2 & 15.9 & 16.9 \\
TIES & & 2.5 & 6.7 & 2.9 & 22.7 & 34.9 & 15.2 & 16.5 \\
DARE & & 2.5 & 4.2 & 1.8 & 20.4 & 29.1 & 13.1 & 13.7 \\
DARE-TIES & & 2.5 & 4.5 & 1.9 & 20.4 & 29.5 & 12.8 & 13.8 \\
\ourmethod(M=1) & & 2.5 &  7.9 & 2.6 & 22.9 & 33.3 & 15.2 & 16.4 \\
\ourmethod(M=40) & & 41.5 & 12.5 & 3.7 & 24.7 & 35.6 & 16.3 & 18.5 \\
\bottomrule
\end{tabular}
}
\end{center}
\end{table*}

\begin{table}[h]
\small
\caption{\textbf{Performance on 40 tasks using Llama-3.2-3B-Instruct LoRA-finetuned} after merging LoRAs across applications, languages and tasks. S represents the percentage of parameters compared to storing all 40 LoRAs.}
\label{table:results_llama_3b_detailed_avg}

\begin{center}
\resizebox{1.15\columnwidth}{!}{%
\hspace*{1em}
\begin{tabular}{lr|rrrrr|r}
\toprule
 Method & S (\%) & \multicolumn{1}{c}{GC} & \multicolumn{1}{c}{SR}  & \multicolumn{1}{c}{TS} & \multicolumn{1}{c}{TA}  & \multicolumn{1}{c|}{QA}  & Avg \\
\midrule
Zero-shot & 0.0 & 10.5 & 3.0 & 19.5 & 28.5 & 14.1 & 15.1 \\
LoRA & 100.0 & 33.7 & 13.4 & 30.9 & 44.0 & 44.0 & 33.2 \\
\hdashline
\tikzmark[xshift=-8pt,yshift=1ex]{x11}TA & 20.0 & 19.5 & 6.7 & 25.2 & 42.2 & 26.7 & 24.1 \\
TIES & 20.0 & 14.9 & 6.0 & 22.9 & 38.7 & 20.1 & 20.5 \\
DARE & 20.0 & 18.7 & 6.9 & 24.5 & 42.9 & 28.2 & 24.2 \\
DARE-TIES & 20.0 & 13.0 & 4.5 & 21.5 & 35.3 & 17.6 & 18.4 \\
\ourmethod (M=1) & 20.0 & 19.3 & 5.7 & 25.6 & 40.4 & 25.8  & 23.3 \\
\ourmethod (M=2) & 28.0 & 21.3 & 10.0 & 26.3 & 42.6 & 32.9  & 26.6 \\
\ourmethod (M=3) & 36.0 & 24.8 & 12.5 & 26.9 & 42.7 & 40.2 & 29.4 \\
\ourmethod (M=4) & 44.0 & 26.7 & 12.6 & 28.5 & 38.8 & 41.8  & 29.7 \\
\tikzmark[xshift=-8pt,yshift=-1ex]{y11}\ourmethod (M=5) & 52.0 & 31.1 & 13.3 & 30.5 & 43.9 & 43.3 & 32.4 \\
\hdashline
\tikzmark[xshift=-8pt,yshift=1ex]{x21}TA & 12.5 & 26.3 & 7.5 & 29.6 & 40.2 & 41.8 & 29.1 \\
TIES & 12.5 & 21.4 & 6.3 & 26.9 & 37.5 & 34.2 & 25.3 \\
DARE & 12.5 & 13.7 & 4.4 & 22.5 & 33.7 & 20.6 & 19.0 \\
DARE-TIES & 12.5 & 15.4 & 5.0 & 24.0 & 35.5 & 25.8 & 21.1 \\
\ourmethod (M=1) & 12.5 & 22.1 & 7.1 & 28.4 & 39.2 & 40.3 & 27.4 \\
\ourmethod (M=2) & 17.5 & 24.5 & 7.8 & 28.8 & 41.4 & 41.1 & 28.7 \\
\ourmethod (M=3) & 22.5 & 27.4 & 8.2 & 29.3 & 42.1 & 41.6 & 29.7 \\
\ourmethod (M=4) & 27.5 & 27.6 & 8.9 & 29.5 & 42.7 & 42.0 & 30.1 \\
\ourmethod (M=5) & 32.5 & 30.6 & 9.4 & 30.4 & 43.2 & 41.1 & 31.0 \\
\ourmethod (M=6) & 37.5 & 29.6 & 10.0 & 30.3 & 43.0 & 42.0 & 31.0 \\
\ourmethod (M=7) & 42.5 & 30.3 & 9.2 & 29.8 & 41.6 & 41.8 & 30.6 \\
\tikzmark[xshift=-8pt,yshift=-1ex]{y21}\ourmethod (M=8) & 47.5 & 31.1 & 12.0 & 30.4 & 43.7 & 43.1 & 32.1 \\
\hdashline
\tikzmark[xshift=-8pt,yshift=1ex]{x31}TA & 2.5 & 17.6 & 5.6 & 24.3 & 39.5 & 22.0 & 21.8 \\
TIES & 2.5 & 13.7 & 4.6 & 22.2 & 35.5 & 17.7 & 18.7 \\
DARE & 2.5 & 11.7 & 3.6 & 20.5 & 32.2 & 15.4 & 16.7 \\
DARE-TIES & 2.5 & 11.6 & 3.6 & 20.4 & 32.3 & 15.2 & 16.6 \\
\ourmethod (M=1) & 2.5 & 18.5 & 5.2 & 24.8 & 39.0 & 22.7 & 22.0 \\
\tikzmark[xshift=-8pt,yshift=-1ex]{y31}\ourmethod (M=40) & 41.5 & 20.4 & 8.0 & 26.7 & 41.8 & 28.9 & 25.2 \\
\bottomrule
\end{tabular}
\drawbrace[brace mirrored, thick]{x11}{y11}
\drawbrace[brace mirrored, thick]{x21}{y21}
\drawbrace[brace mirrored, thick]{x31}{y31}
\annote[left,rotate=90,violet]{brace-4}{applications}
\annote[left,rotate=90,teal]{brace-5}{languages}
\annote[left,rotate=90,purple]{brace-6}{tasks}
}
\end{center}
\end{table}

\begin{table}[!htb]
\caption{\textbf{Performance on 5 English applications using Llama-3.2-1B-Instruct LoRA-finetuned at variable rank.} S represents the percentage of the parameters compared to storing 5 LoRAs. Results are reported per each application separately.}

\small

\label{table:merging_ranks}
\begin{center}
\resizebox{\columnwidth}{!}{%
\begin{tabular}{clc|ccccc|c}
\toprule
Rank & Method & S (\%) & GC & SR & TS & TA & QA & Avg\\
\midrule
& Zero-shot & 0.0 & 13.1 & 5.1 & 23.4 & 27.6 & 15.8 & 17.0 \\
\midrule
\multirow{10}{*}{8} & LoRA & 100.0 & 26.9 & 20.4 & 35.4 & 56.3 & 57.2 & 39.2 \\
\cdashline{2-9}
& TA & 20.0 & 25.4 & 10.6 & 30.1 & 51.1 & 24.6 & 28.4 \\
& TIES & 20.0 & 23.9 & 11.8 & 28.8 & 51.6 & 24.6 & 28.1 \\
& DARE & 20.0 & 17.8 & 6.7 & 25.2 & 35.0 & 18.9 & 20.7 \\
& DARE TIES & 20.0 & 19.9 & 7.5 & 26.0 & 38.6 & 19.8 & 22.4 \\
& \ourmethod (M=1) & 20.0 & 25.1 & 8.5 & 29.9 & 48.0 & 23.8 & 27.1 \\
& \ourmethod (M=2) & 28.0 & 24.6 & 15.5 & 30.0 & 50.6 & 29.5 & 30.0 \\
& \ourmethod (M=3) & 36.0 & 25.5 & 18.7 & 30.4 & 53.5 & 41.4 & 33.9 \\
& \ourmethod (M=4) & 44.0 & 26.3 & 19.1 & 30.1 & 55.9 & 40.2 & 34.3 \\
& \ourmethod (M=5) & 52.0 & 26.5 & 20.0 & 32.8 & 56.7 & 56.4 & 38.5 \\
\midrule
\multirow{10}{*}{16} & LoRA & 100.0 & 32.6 & 21.6 & 36.9 & 57.2 & 60.4 & 41.7 \\
\cdashline{2-9}
& TA & 20.0 & 25.6 & 10.6 & 31.2 & 50.7 & 26.5 & 28.9 \\
& TIES & 20.0 & 24.6 & 13.1 & 29.6 & 52.6 & 26.0 & 29.2 \\
& DARE & 20.0 & 18.7 & 7.0 & 26.1 & 37.8 & 19.6 & 21.8 \\
& DARE TIES & 20.0 & 20.8 & 7.7 & 27.0 & 42.6 & 20.9 & 23.8 \\
& \ourmethod (M=1) & 20.0 & 26.1 & 8.9 & 31.6 & 49.6 & 25.2 & 28.3 \\
& \ourmethod (M=2) & 28.0 & 25.6 & 18.4 & 31.3 & 51.9 & 28.0 & 31.0 \\
& \ourmethod (M=3) & 36.0 & 26.7 & 20.7 & 31.1 & 56.1 & 43.5 & 35.6 \\
& \ourmethod (M=4) & 44.0 & 26.4 & 20.1 & 34.0 & 56.0 & 57.1 & 38.7 \\
& \ourmethod (M=5) & 52.0 & 29.1 & 20.3 & 36.2 & 56.3 & 58.0 & 40.0 \\
\bottomrule
\end{tabular}
}
\end{center}
\end{table}

\begin{table}[!htb]
\caption{\textbf{Performance on 5 English applications using Gemma-2-2B-it LoRA-finetuned.} S represents the percentage of the parameters compared to storing 5 LoRAs. Results are reported per each application separately.}

\small
\label{table:merging_gemma_2b}
\begin{center}
\resizebox{\columnwidth}{!}{%
\begin{tabular}{lc|ccccc|c}
\toprule
Method & S (\%) & GC & SR & TS & TA & QA & Avg\\
\midrule
Zero-shot & 0.0 & 19.4 & 4.3 & 27.4 & 31.0 & 21.7 & 20.8 \\
LoRA & 100.0 & 40.0 & 24.4 & 41.1 & 58.0 & 75.8 & 47.9 \\
\hdashline
TA & 20.0 & 29.5 & 14.9 & 34.9 & 51.4 & 57.9 & 37.7 \\
TIES & 20.0 & 28.6 & 14.9 & 33.0 & 51.4 & 47.7 & 35.1 \\
DARE & 20.0 & 23.1 & 4.5 & 28.9 & 38.6 & 27.8 & 24.6 \\
DARE TIES & 20.0 & 24.0 & 5.0 & 29.0 & 41.1 & 29.5 & 25.7 \\
\ourmethod (M=1) & 20.0 & 29.3 & 12.2 & 35.9 & 49.4 & 53.5 & 36.0 \\
\ourmethod (M=2) & 28.0 & 28.7 & 20.7 & 36.1 & 53.6 & 62.5 & 40.3 \\
\ourmethod (M=3) & 36.0 & 28.5 & 23.1 & 37.2 & 57.1 & 67.3 & 42.6 \\
\ourmethod (M=4) & 44.0 & 26.5 & 23.2 & 39.2 & 56.5 & 72.8 & 43.7 \\
\ourmethod (M=5) & 52.0 & 39.2 & 23.7 & 40.8 & 58.1 & 75.8 & 47.5 \\
\bottomrule
\end{tabular}
}
\end{center}
\end{table}
\begin{table}[!htb]
\caption{\textbf{Performance on 5 English applications using Phi-3.5-mini-instruct LoRA-finetuned.} S represents the percentage of the parameters compared to storing 5 LoRAs. Results are reported per each application separately.}
\small

\label{table:merging_phi_3_5b}
\begin{center}
\resizebox{\columnwidth}{!}{%
\begin{tabular}{lc|ccccc|c}
\toprule
Method & S (\%) & GC & SR & TS & TA & QA & Avg\\
\midrule
Zero-shot & 0.0 & 19.1 & 2.6 & 23.6 & 21.5 & 7.5 & 14.9 \\
LoRA & 100.0 & 33.3 & 25.1 & 39.9 & 58.0 & 71.0 & 45.4 \\
\hdashline
TA & 20.0 & 26.8 & 10.3 & 33.1 & 49.4 & 47.0 & 33.3 \\
TIES & 20.0 & 26.7 & 12.9 & 32.1 & 52.3 & 46.4 & 34.1 \\
DARE & 20.0 & 22.3 & 3.7 & 25.9 & 38.1 & 10.8 & 20.1 \\
DARE TIES & 20.0 & 22.3 & 3.7 & 25.9 & 38.1 & 10.8 & 20.1 \\
\ourmethod (M=1) & 20.0 & 26.7 & 7.4 & 33.7 & 45.6 & 39.5 & 30.6 \\
\ourmethod (M=2) & 28.0 & 27.4 & 18.4 & 34.7 & 52.2 & 71.1 & 40.8 \\
\ourmethod (M=3) & 36.0 & 28.0 & 23.3 & 36.7 & 57.5 & 71.0 & 43.3 \\
\ourmethod (M=4) & 44.0 & 29.7 & 21.0 & 36.8 & 56.8 & 67.7 & 42.4 \\
\ourmethod (M=5) & 52.0 & 32.7 & 25.0 & 40.0 & 57.7 & 70.7 & 45.2 \\
\bottomrule
\end{tabular}
}
\end{center}
\end{table}

\begin{table*}[!htb]
\setlength{\tabcolsep}{3pt}
\caption{\textbf{Comparison of the methods in terms of runtime and GPU memory.} The results indicate that HydraOpt introduces additional runtime; however, the merging operation is still reasonably fast with relatively small GPU memory overhead.}
\label{table:runtime_memory}
\begin{center}
\resizebox{\textwidth}{!}{%
\begin{tabular}{crrrrrrrrr}
\toprule
Metric & TA & TIES & DARE & DARE-TIES & HydraOpt(M=1) & HydraOpt(M=2) & HydraOpt(M=3) & HydraOpt(M=4) & HydraOpt(M=5) \\
\midrule
Runtime (mins) & 0.7 & 0.6 & 0.7 & 0.7 & 10.6 & 13.9 & 17.2 & 20.6 & 8.6 \\
GPU (GB) & 19.6 & 19.6 & 19.6 & 19.6 & 20.5 & 20.7 & 21.0 & 21.3 & 20.7 \\
\bottomrule
\end{tabular}
}
\end{center}
\end{table*}

\end{document}